\documentclass[lettersize,journal]{IEEEtran}
\usepackage{amsmath,amsfonts}
\usepackage{amssymb}
\usepackage{graphicx}
\usepackage{algorithm}
\usepackage{array}
\usepackage[caption=false,font=normalsize,labelfont=sf,textfont=sf]{subfig}
\usepackage{textcomp}
\usepackage{stfloats}
\usepackage{url}
\usepackage{verbatim}
\usepackage{graphicx}
\usepackage{cite}
\usepackage{multirow}
\usepackage{multicol}
\usepackage{colortbl}
\usepackage{pifont}
\usepackage{algpseudocode}

\algrenewcommand\textproc{}
\begin{document}

\title{Trainable Pointwise Decoder Module for Point Cloud Segmentation}

\author{Bike Chen, Chen Gong, Antti Tikanmäki, Juha Röning,
	% <-this % stops a space
	%\thanks{Bike Chen, Antti Tikanmäki, and Juha Röning are with the Biomimetics and Intelligent Systems Group, University of Oulu, Finland. Chen Gong is with LEAP Group, Nanjing University of Science and Technology, China. (e-mail: \{bike.chen, antti.tikanmaki, juha.roning\}@oulu.fi, chen.gong@njust.edu.cn) \par \emph{Corresponding author: Juha Röning.}}% <-this % stops a space
	%\thanks{Manuscript received April 19, 2021; revised August 16, 2021.}
}

% The paper headers
\markboth{Journal of \LaTeX\ Class Files,~Vol.~14, No.~8, August~2021}%
{Shell \MakeLowercase{\textit{et al.}}: A Sample Article Using IEEEtran.cls for IEEE Journals}

%\IEEEpubid{0000--0000/00\$00.00~\copyright~2021 IEEE}
% Remember, if you use this you must call \IEEEpubidadjcol in the second
% column for its text to clear the IEEEpubid mark.

\maketitle

\begin{abstract}
Point cloud segmentation (PCS) aims to make per-point predictions and enables robots and autonomous driving cars to understand the environment. The range image is a dense representation of a large-scale outdoor point cloud, and segmentation models built upon the image commonly execute efficiently. However, the projection of the point cloud onto the range image inevitably leads to dropping points because, at each image coordinate, only one point is kept despite multiple points being projected onto the same location. More importantly, it is challenging to assign correct predictions to the dropped points that belong to the classes different from the kept point class. Besides, existing post-processing methods, such as $K$-nearest neighbor ($K$NN) search and kernel point convolution (KPConv), cannot be trained with the models in an end-to-end manner or cannot process varying-density outdoor point clouds well, thereby enabling the models to achieve sub-optimal performance. To alleviate this problem, we propose a trainable pointwise decoder module (PDM) as the post-processing approach, which gathers weighted features from the neighbors and then makes the final prediction for the query point. In addition, we introduce a virtual range image-guided copy-rotate-paste (VRCrop) strategy in data augmentation. VRCrop constrains the total number of points and eliminates undesirable artifacts in the augmented point cloud. With PDM and VRCrop, existing range image-based segmentation models consistently perform better than their counterparts on the SemanticKITTI, SemanticPOSS, and nuScenes datasets.
\end{abstract}


\section{Introduction} \label{sec:intro}
% Step 1: Background.
\IEEEPARstart{P}{oint} cloud segmentation (PCS) aims to assign a label to each point in the point cloud. It plays an important role in robots and self-driving cars. Specifically, the outputs of PCS enable robots and autonomous driving cars to understand the physical environments~\cite{semantickitti_2019_behley,semanticposs_2020,nuscenes_panoptic,domainAdpt2024}. Also, PCS results can be used to make a high-definition semantic map~\cite{sa_loam_2021, suma++_2019} for robot navigation.  

To efficiently parse the sparse, unordered, and irregular points in a large-scale outdoor point cloud~\cite{semantickitti_2019_behley,semanticposs_2020,nuscenes_panoptic}, the structured and compact range image representation is adopted to serve as a proxy. Corresponding models usually require lower computational cost than point-based~\cite{pointnet++_2017,randla_2020} and voxel-based models~\cite{spvnas_2020,cylindrical_2021,spherical_transformer_2023}.

\begin{figure}[t]
	\centering
	\includegraphics[width=0.86\columnwidth]{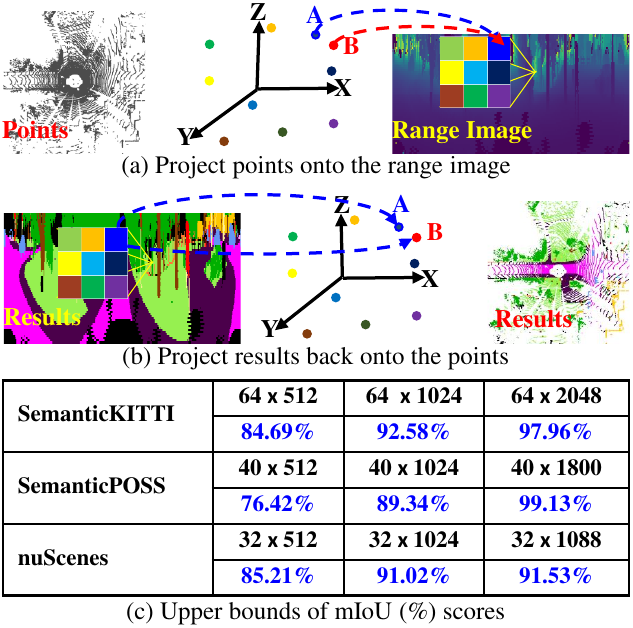}
	\caption{(a) When points are projected onto the range image, some points might be projected onto the same location due to the discretization. For example, points \textcolor{blue}{A} and \textcolor{red}{B} are mapped to the same position (\textit{i.e.}, the \textcolor{blue}{blue grid}) in the range image. (b) Correspondingly, when per-pixel predictions are projected back onto the points, the points belonging to different classes might be assigned the same label. For instance, the \textcolor{blue}{A} and \textcolor{red}{B} belong to different classes but get the same prediction from the range image result. This leads to inferior per-point segmentation performance. (c) Theoretical upper bounds of mIoU ($\%$) scores under various sizes of range images on SemanticKITTI~\cite{semantickitti_2019_behley}, SemanticPOSS~\cite{semanticposs_2020} and nuScenes~\cite{nuscenes_panoptic} datasets. Specifically, under the sizes of $64\times2048$, $40\times1800$, and $32\times1088$, a range image-based model achieves at most 97.96\%, 99.13\%, and 91.53\% mIoU scores on the three datasets, respectively.}
	\label{fig:upper_bound_miou}  
\end{figure}

% Step 2: What is the motivation of your work? What is the significance of your work?
However, projecting a point cloud onto the range image inevitably results in the loss of information~\cite{rangenet++,scan_based_projection,filling_missing2024} due to the discretization\footnote{Scan unfolding++~\cite{filling_missing2024} is adopted to prepare the range image}, thereby degenerating the performance of range image-based models. Specifically, the azimuth angles of the points are strongly related to the horizontal coordinates of the points in the range image. After the discretization of the azimuth angles, adjacent points having similar angles might have the same horizontal coordinate. More importantly, when the adjacent points are projected onto the same image coordinate, only one point is kept at each image position and the rest are dropped (see Fig.~\ref{fig:upper_bound_miou}(a)). This leads to the loss of information. Moreover, when the per-pixel predictions of the range image are projected back onto the points, the dropped points are assigned the same prediction as the kept point (see Fig.~\ref{fig:upper_bound_miou}(b)). If the dropped points belong to different classes from the kept point, the final segmentation performance is decreased. Theoretically, all range image-based PCS models have the upper bounds of mIoU (\%) scores on various datasets (see Fig.~\ref{fig:upper_bound_miou}(c)). Therefore, an effective and efficient post-processing module is required as a supplement to the range image-based PCS models to break the theoretical upper bounds of performance. 

% Step 2-1: Why can others not do this?
Existing $K$-nearest neighbor ($K$NN) search~\cite{rangenet++} and nearest label assignment (NLA)~\cite{fidnet_2021} cope with the problem above by utilizing the predictions of the nearest neighbors for the query point within a local area. However, the range image-based models cannot be trained with $K$NN and NLA in an end-to-end manner, thereby achieving sub-optimal performance. Besides, KPConv~\cite{kpconv_2019,kprnet_2020,rangevit_2023} can be trained with the range image-based models together, but KPConv requires high computational cost and cannot process varying-density outdoor point clouds well. Hence, the performance gain of the models with KPConv is limited.  

In addition, to the best of our knowledge, no copy-rotate-paste or copy-paste strategy in data augmentation for the problem of class imbalance is suitable for training the image-point fused models. Existing copy-rotate-paste or copy-paste strategy~\cite{fidnet_2021,cenet_2022,polarmix_2022} directly appends the points of rare instances to the current point cloud. However, the large-scale outdoor point cloud already contains a substantial number of points (\textit{i.e.}, about 130,000 points~\cite{semantickitti_2019_behley}). After data augmentation, the total number of points during training is further increased, and processing all points at once will be impractical with a limited computational resource. Besides, the conventional strategy introduces undesirable artifacts which bring difficulty in training image-point fused models and decrease the performance. 

\begin{figure}[t]
	\centering
	\includegraphics[width=1.0\columnwidth]{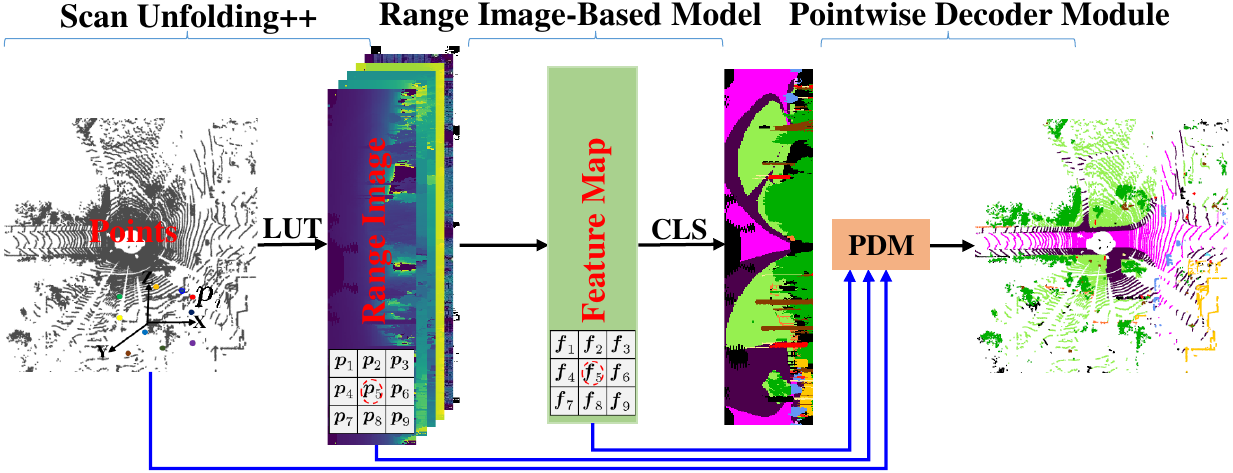}
	\caption{Overview of the framework. Scan unfolding++~\cite{filling_missing2024} is used to project points onto the range image. Then, the range image goes through a range image-based model to produce the feature map and make per-pixel predictions. Finally, the proposed trainable pointwise decoder module (PDM) is utilized to make per-point predictions.}
	\label{fig:overview}  
\end{figure}

% Step 3: Therefore, you propose a method which can solve previous problems (Why).
To cope with the problems above, we propose an effective and efficient trainable pointwise decoder module (PDM), which can be trained with existing range image-based models in an end-to-end fashion (see Fig.~\ref{fig:overview}). PDM can be split into two parts: range image-guided $K$NN search and a local feature extraction module. The range image-guided $K$NN search is inspired by the $K$NN in RangeNet++~\cite{rangenet++}. $K$ nearest neighbor points of a query point in the 3D space are the $K$ nearest neighbor pixels of the center pixel within a local window on the range image. However, different from that in RangeNet++, the range image for our $K$NN search is prepared by scan unfolding++~\cite{filling_missing2024} instead of spherical projection. This is beneficial to train the models. Besides, our method discards the operations of Gaussian weighting and range cutoff~\cite{fidnet_2021}, thereby being efficient (see Sec.~\ref{sec:knn_search}).

The local feature extraction module in PDM is similar to the point transformer layer in the work~\cite{pointtrans2021}. Attentive weights are computed based on relative positions and edge features. Then, the weights are applied to fused features to produce the features of the query point. Finally, a classifier is used to make a prediction. Different from the point transformer layer, the nearest neighbor points in this paper are provided by the range image-guided $K$NN search. Besides, the linear function on features is discarded because the local feature extraction module aims to refine the ``final" features and should be simple and efficient. Moreover, a classifier is added to output the per-point prediction (see Sec.~\ref{sec:local_feature_extra}).

In addition, we propose a virtual range image-guided copy-rotate-paste (VRCrop) strategy in data augmentation for training range image-point fused models (\textit{e.g.}, RangeNet+PDM). All points in a point cloud are projected onto a virtual range image, and then each point has a vertical and horizontal coordinate. Like the copy-paste operation for color images, VRCrop pastes the points of the rare instances to the current point cloud in which the points with the same coordinates as the pasted points are deleted in advance. With VRCrop, the maximal number of points in the point cloud after augmentation is limited. More importantly, the problem of existing undesirable artifacts is alleviated (see Sec.~\ref{sec:vrcrop_strategy}). 

Extensive experiments conducted on SemanticKITTI \cite{semantickitti_2019_behley}, SemanticPOSS \cite{semanticposs_2020}, and nuScenes~\cite{nuscenes_panoptic} datasets show that by employing the proposed PDM and VRCrop, existing range image-based segmentation models achieve better performance than their counterparts. Our contributions are summarized as follows:

\begin{enumerate}
	\item We propose a trainable pointwise decoder module (PDM) to improve the performance of existing range image-based models. For a query point, PDM uses range image-guided $K$NN search to efficiently find its $K$ nearest neighbor points, and adopt a local feature extraction module to extract features and make an effective prediction.
	
	\item We introduce a virtual range image-guided copy-rotate-paste (VRCrop) strategy instead of the conventional copy-rotate-past or copy-paste operation in point cloud data augmentation. With VRCrop, the point cloud after data augmentation has a limited number of points, thereby requiring less computational cost during training image-point fused models. Besides, the problem of occurring undesirable artifacts in the augmented point cloud is alleviated.   
\end{enumerate}

\section{Related Work}\label{sec:related_works}
In this section, previous works associated with this paper are briefly reviewed. 

\subsection{Range Image-based Point Cloud Segmentation} 
The development of different backbones plays a crucial role in the point cloud segmentation (PCS) task. For example, SqueezeSeg~\cite{squeezeseg} pioneers the range image-based approach in the PCS task and adopts SqueezeNet~\cite{squeezenet_2016} as its backbone. Subsequently, RangeNet++~\cite{rangenet++} utilizes revised DarkNet~\cite{yolov3_2018} as its backbone. Both FIDNet~\cite{fidnet_2021} and CENet~\cite{cenet_2022} use ResNet34~\cite{resnet_2016} as their backbones. FMVNet and Fast FMVNet~\cite{filling_missing2024} adopt modified ConvNeXt~\cite{convnext2022} as the backbone. These are convolution-based segmentation models, and they are usually efficient. In addition, RangeViT~\cite{rangevit_2023} choose ViT~\cite{vit_iclr_2021} as the backbone, and RangeFormer~\cite{rangeformer_2023} is built on the SegFormer-like~\cite{segformer2021} architecture. These are transformer-based PCS models, which commonly achieve better performance than convolution-based models, but run at relatively slow speeds. In this paper, we do not focus on building a backbone. Instead, we introduce a trainable pointwise decoder module integrated into existing range image-based models. The improved models achieve better performance than their counterparts in terms of mIoU scores.

\subsection{Post-processing Methods}
Due to the loss of information caused by the discretization in the range image preparation, post-processing methods over per-point predictions are required to break the theoretical upper bounds of segmentation performance. $K$-nearest neighbor ($K$NN) search~\cite{rangenet++,cenet_2022} utilizes the predictions of $K$ nearest neighbor points to vote for the prediction of the query point. Nearest label assignment (NLA)~\cite{fidnet_2021} directly assigns one nearest neighbor point prediction to the query point. Both $K$NN and NLA can improve the performance of range image-based models. However, the $K$NN and NLA cannot be trained with the range image models together. Hence, the final models can only achieve sub-optimal performance. In addition, RangePost~\cite{rangeformer_2023} first splits a whole point cloud into several sub-point clouds, then gets the predictions of each subset, and finally merges all predictions. This approach can avoid the loss of information but decreases the model inference speed. Besides, KPConv~\cite{kpconv_2019} is used as the post-processing module in the works~\cite{kprnet_2020,rangevit_2023}. KPConv can be trained with range image-based models in an end-to-end manner and refines the final predictions. However, KPConv has a high computational cost and cannot process varying-density outdoor point clouds well. Similarly, in this paper, we propose a trainable pointwise decoder module (PDM), which can be trained with the range image-based models in an end-to-end fashion and copes with varying-density point clouds well. The models with PDM obtain better performance than their counterparts.

\subsection{Data Augmentation Techniques}
Data augmentation techniques make point clouds diverse and significantly increase the performance of point cloud segmentation models. The commonly used techniques~\cite{data_aug2024} are random translation, random rotation, random flipping, random scaling, random dropping, and random jittering. Besides, for LiDAR data, PolarMix~\cite{polarmix_2022} swaps two subsets from two point clouds along the horizontal direction (\textit{i.e.}, scene-level mixing). Similarly, LaserMix~\cite{lasermix_2023,rangeformer_2023} swaps the subsets from the two point clouds along the vertical direction (\textit{i.e.}, RangeMix). These data augmentation techniques do not increase the total number of points and, hence, do not raise the computational cost. In addition, the works~\cite{fidnet_2021,cenet_2022,polarmix_2022,lasermix_2023,rangeformer_2023} utilize the copy-rotate-paste or copy-paste augmentation (\textit{i.e.}, instance-level mixing in PolarMix; RangePaste in RangeFormer) to alleviate the class-imbalance problem in datasets. However, RangePaste is only applied to range images, so it is not suitable for image-point fused models. The rest directly appends the points of rare instances from a random point cloud to the current point cloud. This leads to a high computational cost due to an increased number of points in the current point cloud. Moreover, this introduces undesirable artifacts, thereby decreasing the segmentation performance of the image-point fused models. In this paper, we propose a virtual range image-guided copy-rotate-paste (VRCrop) strategy to improve the copy-rotate-paste augmentation. With VRCrop, the maximal number of points in the point cloud after data augmentation is limited. More importantly, the problem of occurring unwanted artifacts is alleviated.

\subsection{Feature Extraction Modules}
For the point cloud, a feature extraction module should be specially designed to extract semantic features over points. PointNet~\cite{pointnet2017} is a pioneer which uses fully connected layers to extract features from each point and then adopts a global pooling operation to aggregate the global information. However, PointNet fails to incorporate a local feature extraction component into its architecture, thereby degenerating performance. To overcome the problem in PointNet, a wide range of approaches~\cite{pointnet++_2017,pcnn2018,dgcnn2019,closer3d2020,pointtrans2021,neigh_agg2024} include various local feature extraction modules in their architectures. Typically, within a local region, PointNet++~\cite{pointnet++_2017} applies several MLP layers on the concatenation of point features and relative positions, and then uses a max pooling operation to aggregate the transformed features for the query point. By contrast, DGCNN~\cite{dgcnn2019} adopts the concatenation of point features and edge features as the input of MLP layers. PCCN~\cite{pcnn2018} applies the MLP layers on relative positions to compute weights, which are subsequently adopted to aggregate features from neighbor points. Point transformer~\cite{pointtrans2021} employs the self-attention technique to relative positions and edge features to extract local features for the query point. All these local feature extraction modules can improve performance. However, no detailed comparison results among them have been found when they are combined with range image-based models. In this paper, we provide many options and show detailed comparison results. Also, considering the speed-accuracy trade-off, we choose the best one integrated into the proposed trainable pointwise decoder module.

\section{Trainable Pointwise Decoder Module}\label{sec:trainable_pdm}
In this section, we detail the proposed trainable pointwise decoder module (PDM). In PDM, two parts, namely the range image-guided $K$NN search and the local feature extraction module, are described sequentially. Besides, for ease of description, we set the window size to $3\times3$ and top-$K$ to 5 in $K$NN. 

\begin{figure}[t]
	\centering
	\includegraphics[width=1.0\columnwidth]{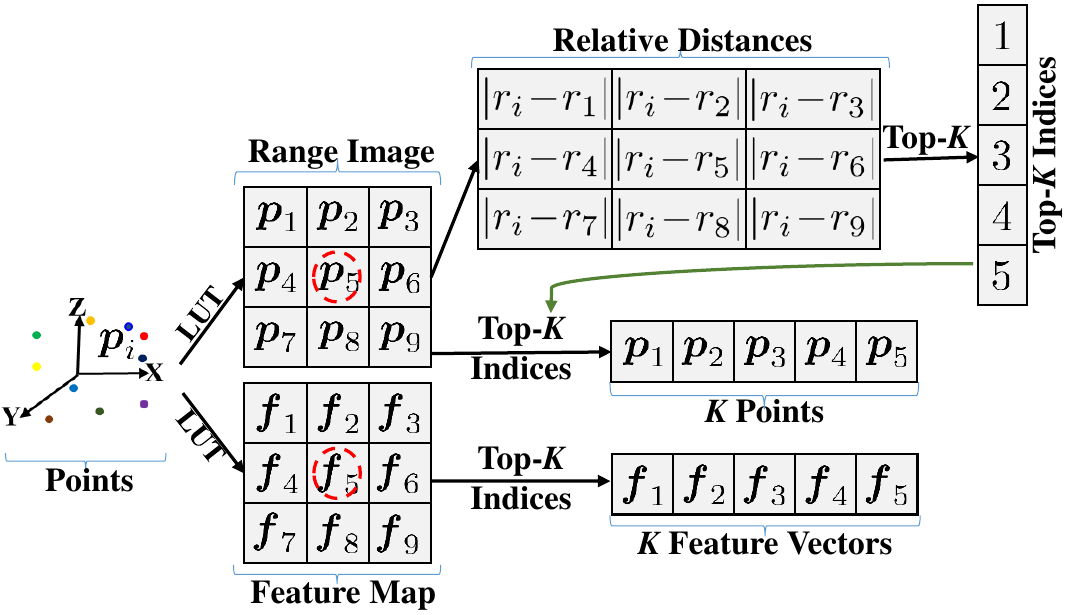}
	\caption{Range image-guided $K$NN search in PDM. According to the look-up table (LUT), a point $\boldsymbol{p}_i$ can first find its $\left(u, v\right)$ coordinate in ``Range Image" and ``Feature Map" (indicated by the dash red circles). Then, the point can efficiently search the neighbor points $\left\{\boldsymbol{p}_1, \boldsymbol{p}_2, \dots, \boldsymbol{p}_9\right\}$ and corresponding features $\left\{\boldsymbol{f}_1, \boldsymbol{f}_2, \dots, \boldsymbol{f}_9\right\}$. Next, based on ``Relative Distances", ``Top-$K$ Indices" can be calculated. Finally, the indices are employed to choose $K$ nearest neighbor points $\left\{\boldsymbol{p}_1, \boldsymbol{p}_2, \dots, \boldsymbol{p}_5\right\}$ and corresponding feature vectors $\left\{\boldsymbol{f}_1, \boldsymbol{f}_2, \dots, \boldsymbol{f}_5\right\}$.}
	\label{fig:pdm_knn}  
\end{figure}

\subsection{Range Image-guided $K$NN Search}\label{sec:knn_search}
The component is to efficiently find $K$ nearest neighbor points and corresponding feature vectors for the following local feature extraction module (see Fig.~\ref{fig:pdm_knn}). 

This part includes the following five steps:

\textbf{Step 1:} For the point $\boldsymbol{p}_i$, the $\left(u, v\right)$ coordinate can be obtained based on the look-up table (LUT)~\cite{filling_missing2024} where the indices of all points in a scan are associated with the corresponding $\left(\boldsymbol{u}, \boldsymbol{v}\right)$ coordinates in the range image.  

\textbf{Step 2:} According to the $\left(u, v\right)$ coordinate, the neighbor points and corresponding feature vectors can be easily searched. For example, in Fig.~\ref{fig:pdm_knn}, the $\left(u, v\right)$ positions in ``Range Image" and ``Feature Map" are indicated by dashed red circles. The neighbor points $\left\{\boldsymbol{p}_1, \boldsymbol{p}_2, \dots, \boldsymbol{p}_9\right\}$ and corresponding feature vectors $\left\{\boldsymbol{f}_1, \boldsymbol{f}_2, \dots, \boldsymbol{f}_9\right\}$ within the $3\times3$ window can be efficiently retrieved.

\textbf{Step 3:} Relative distances between the point $\boldsymbol{p}_i$ and its neighbor points are calculated. For instance, in Fig.~\ref{fig:pdm_knn}, the ``Relative Distances" $\left\{|r_i - r_1|, |r_i - r_2|, \dots, |r_i - r_9|\right\}$ are computed.

\textbf{Step 4:} Based on the relative distances, the indices of the top-$K$ nearest neighbor points are obtained. For example, in Fig.~\ref{fig:pdm_knn}, the indices of the top-$K$ small absolute values over the relative distances are selected. For ease of description, we assume that the ``Top-$K$ Indices" are $\left\{1, 2, 3, 4, 5\right\}$. Note that the small relative distance between two points within a local window in the range image means the short distance between the points in the 3D space~\cite{rangenet++}.

\textbf{Step 5:} The top-$K$ indices are applied on the range image and feature map to select the $K$ nearest neighbor points and corresponding feature vectors. For example, in Fig.~\ref{fig:pdm_knn}, with ``Top-$K$ Indices", the $\left\{\boldsymbol{p}_1, \boldsymbol{p}_2, \dots, \boldsymbol{p}_5\right\}$ and the corresponding $\left\{\boldsymbol{f}_1, \boldsymbol{f}_2, \dots, \boldsymbol{f}_5\right\}$ are selected for the following local feature extraction module.

\begin{figure}[t]
	\centering
	\includegraphics[width=0.8\columnwidth]{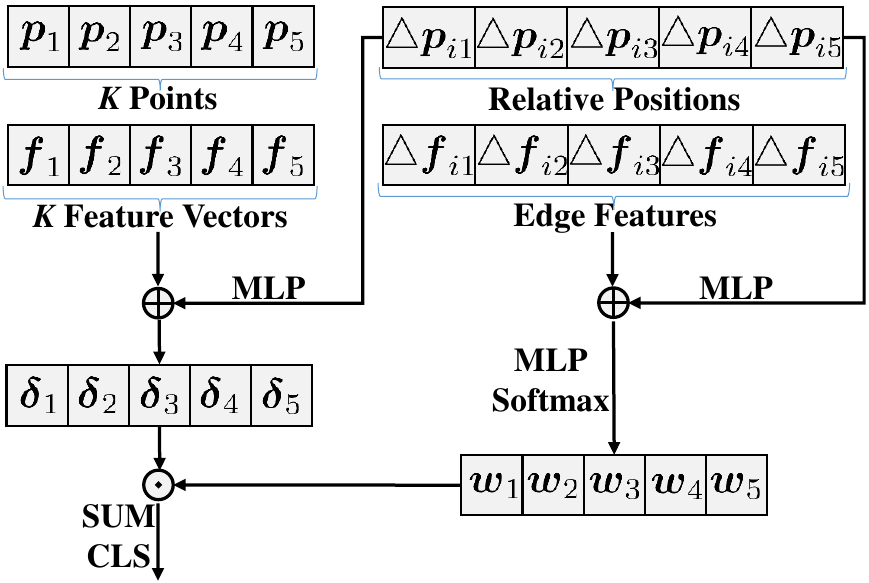}
	\caption{Local feature extraction module in PDM. ``Relative Positions" $\left\{\triangle\boldsymbol{p}_{i1}, \triangle\boldsymbol{p}_{i2}, \dots, \triangle\boldsymbol{p}_{i5}\right\}$ and ``Edge Features" $\left\{\triangle\boldsymbol{f}_{i1}, \triangle\boldsymbol{f}_{i2}, \dots, \triangle\boldsymbol{f}_{i5}\right\}$ are used to calculate attentive weights $\left\{\boldsymbol{w}_1, \boldsymbol{w}_2, \dots, \boldsymbol{w}5\right\}$. ``Feature Vectors" $\left\{\boldsymbol{f}_1, \boldsymbol{f}_2, \dots, \boldsymbol{f}_5\right\}$ and ``Relative Positions" are fused to generate the features $\left\{\boldsymbol{\delta}_1, \boldsymbol{\delta}_2, \dots, \boldsymbol{\delta}_5\right\}$. Finally, the summation (SUM) operation is adopted to aggregate the weighted features, and a classifier is utilized to make per-point predictions.}
	\label{fig:pdm_attn_weight}  
\end{figure}

\subsection{Local Feature Extraction Module}\label{sec:local_feature_extra}
The module aims to gather features for the query point over the $K$ nearest neighbor points and then make a prediction for the query point (see Fig.~\ref{fig:pdm_attn_weight}). This module can refine the per-point predictions of the models, especially for the dropped points in the range image.

The module contains the following four steps:

\textbf{Step 1:} Relative positions and edge features are computed by subtracting the position and feature vector of the query point from the neighbor points. For example, in Fig.~\ref{fig:pdm_attn_weight}, ``Relative Positions" $\triangle\boldsymbol{p}_{ij} = |\boldsymbol{p}_{j} - \boldsymbol{p}_{i}|, j \in \mathcal{N}\left(i\right)$ and ``Edge Features" $\triangle\boldsymbol{f}_{ij} = \boldsymbol{f}_{j} - \boldsymbol{f}_{i}, j \in \mathcal{N}\left(i\right)$ are obtained. Here, the $\mathcal{N}\left(i\right)$ means the indices of the neighbor points.

\textbf{Step 2:} Attentive weights are calculated by the expression $\boldsymbol{w}_j =\text{Softmax}\left( \text{MLP}\left(\triangle\boldsymbol{f}_{ij} + \text{MLP}\left(\triangle\boldsymbol{p}_{ij}\right) \right)\right)$. Here, MLP is comprised of two linear layers, a batch normalization layer, and an activation layer. The expression indicates that the weights are only derived from the position and feature relationships.  

\textbf{Step 3:} Fused features are computed by the formula $\boldsymbol{\delta}_j = \boldsymbol{f}_j + \text{MLP}\left(\triangle\boldsymbol{p}_{ij}\right)$. The relative position information is crucial for the final per-point prediction, so the fused features contain the relative information $\text{MLP}\left(\triangle\boldsymbol{p}_{ij}\right)$.

\textbf{Step 4:} A summation operation (SUM) is utilized to aggregate all features for the query point $\boldsymbol{p}_i$. This is expressed by the formula $\boldsymbol{o}_i = \text{SUM}\left\{\boldsymbol{w}_j \odot \boldsymbol{\delta}_j | j \in \mathcal{N}(i) \right\}$. Finally, a classifier is adopted to make the final prediction. The classifier contains two linear layers, a batch normalization layer, and an activation layer.

\begin{algorithm}[t]
	\caption{Scan Unfolding++~\cite{filling_missing2024}.}
	\label{alg:scan_unfolding++}
	\small
	\begin{algorithmic}[1]
		\Require $N$ points $\boldsymbol{P}$ in a point cloud; The ring numbers $\boldsymbol{d}$ for all points in the point cloud; The width of the range image $W$.
		
		\Ensure The $\left(\boldsymbol{u}, \boldsymbol{v}\right)$ coordinates for all points in the range image.
		
		\vspace{2ex}
		
		\State $\boldsymbol{x} = \boldsymbol{P}[:, 0]$, $\boldsymbol{y} = \boldsymbol{P}[:, 1]$. 
		
		\State $\boldsymbol{\theta} = \text{arctan}(\boldsymbol{y} / \boldsymbol{x}) \times 180 / \pi$.
		
		\State $\boldsymbol{m} = \boldsymbol{\theta} < 0$.
		
		\State $\boldsymbol{\theta}[\boldsymbol{m}] = \boldsymbol{\theta}[\boldsymbol{m}] + 360$.
		
		\State $\boldsymbol{u} = \boldsymbol{\theta} / 360 \times W$.
		
		\State $\boldsymbol{v} = \boldsymbol{d}$.
		
	\end{algorithmic}
\end{algorithm}

\begin{figure}[!htp]
	\centering
	\includegraphics[width=0.75\columnwidth]{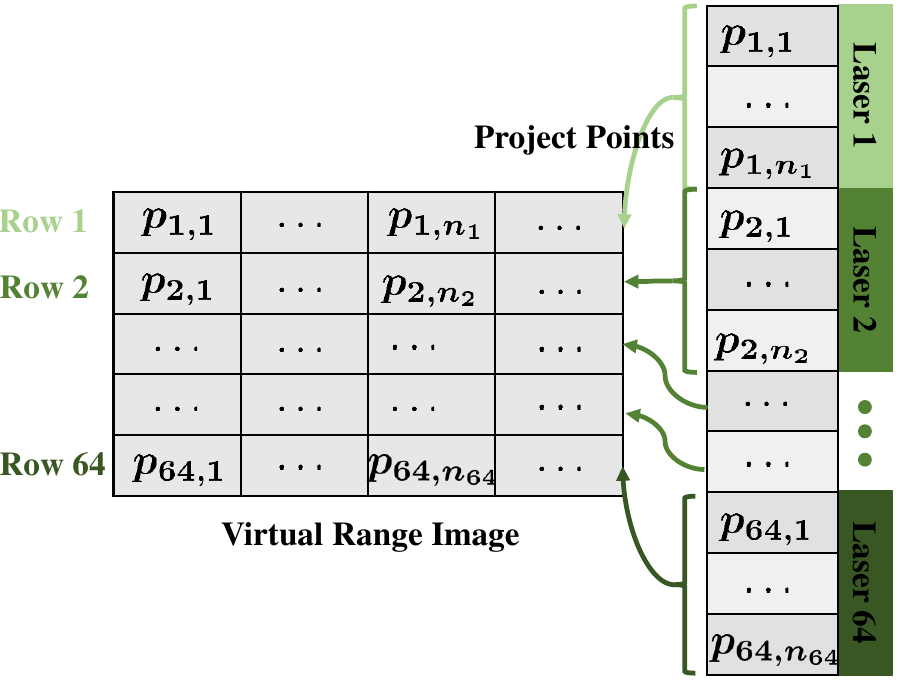}
	\caption{Projection of points onto the virtual range image. The points from the same laser are sequentially projected onto the same scan line. If two or more projected points occupy one virtual range image position, all points are kept. No points are dropped.}
	\label{fig:project_points_range_image}
\end{figure}

\section{Virtual Range Image-guided Copy-rotate-paste Strategy}\label{sec:vrcrop_strategy}
In this section, we first introduce how to produce a virtual range image. Then, we detail the proposed virtual range image-guided copy-rotate-paste (VRCrop) strategy. The utilization of VRCrop avoids a considerable increase in the number of points in the point cloud after data augmentation. Moreover, the use of VRCrop alleviates the problem of undesirable artifacts in the augmented point cloud.

\subsection{Virtual Range Image Preparation}
The virtual range image should satisfy two requirements: (1) No points are dropped in the virtual range image to avoid the loss of information before the copy-rotate-paste operation; (2) Every point in the virtual image should have the row $v$ and column $u$ coordinate. This helps to detect conflicted points when copying instances from a random point cloud to the current point cloud.

To meet the first requirement, a sufficiently large virtual range image should be created to store all projected points. The image size is associated with sensor properties. For example, the virtual range image size for the SemanticKITTI dataset~\cite{semantickitti_2019_behley} should be set to $64\times2180$, because the data is collected by a 64-beam LiDAR sensor~\cite{hdl_64e_s2_manual}, and there are at most 2180 points per scan line~\cite{filling_missing2024}. 

To fulfil the second requirement, the projection method, scan unfolding++~\cite{filling_missing2024}, is adopted to calculate $\left(u, v\right)$ coordinate for each point. Scan unfolding++ is described by Alg.~\ref{alg:scan_unfolding++}.

First, the azimuth angles for the points are calculated (Lines 1 \& 2). Then, the negative azimuth angles are converted to the positive values (Lines 3 \& 4). This makes all azimuth angles within $\left[0, 360\right]$. Finally, horizontal coordinates for the points are computed by the expression in Line 5, and the vertical coordinates are the ring numbers. 

According to the computed $\left(\boldsymbol{u}, \boldsymbol{v}\right)$ coordinates, all points from the same laser are sequentially projected onto the corresponding row of the virtual range image (see Fig.~\ref{fig:project_points_range_image}). Besides, note that if more points occupy the same position in the virtual range image, all of them are kept. This is different from the preparation of the range image.

\subsection{VRCrop Strategy}
VRCrop strategy includes four steps: copying points from a random point cloud, rotating the copied points, detecting and deleting the conflicted points in the current point cloud, and pasting the rotated points to the current point cloud. For ease of description, we assume that the current scan is ``Point Cloud A" and a random scan is ``Point Cloud B". Each point in the virtual range image is expressed by $\boldsymbol{p}_{ij}$ where $i$ corresponds to the vertical coordinate $v$ (or the laser ring number), and $j$ indicates the horizontal coordinate $u$. A rare instance is indicated by the points $\left\{\boldsymbol{p}_{22}, \boldsymbol{p}_{23}, \boldsymbol{p}_{32}, \boldsymbol{p}_{33}\right\}$. Besides, we assume that each point contains a label. 

\begin{figure}[t]
	\centering
	\includegraphics[width=0.8\columnwidth]{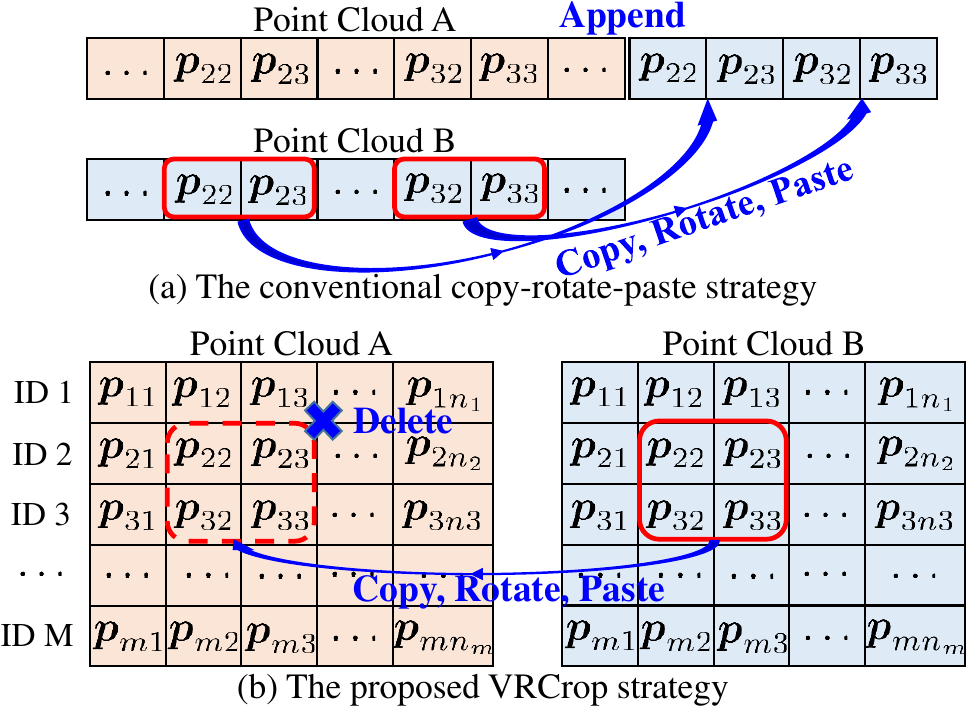}
	\caption{(a) The conventional copy-rotate-paste operation. The points  $\left\{\boldsymbol{p}_{22}, \boldsymbol{p}_{23}, \boldsymbol{p}_{32}, \boldsymbol{p}_{33}\right\}$ of the rare instance from ``Point Cloud B" are copied, rotated, and pasted to ``point cloud A". Note that the rotated points are directly appended to ``Point Cloud A". (b) The proposed VRCrop strategy. First, all points are projected onto the virtual range image. Second, the points $\left\{\boldsymbol{p}_{22}, \boldsymbol{p}_{23}, \boldsymbol{p}_{32}, \boldsymbol{p}_{33}\right\}$ of the rare instance are copied from ``Point Cloud B". Third, the points are rotated and have updated $\left(\boldsymbol{u}, \boldsymbol{v}\right)$ coordinates. Fourth, VRCrop strategy detects and deletes the conflicted points in ``Point Cloud A" with the same $\left(\boldsymbol{u}, \boldsymbol{v}\right)$ coordinates. Fifth, the rotated points are pasted to ``Point Cloud A".}
	\label{fig:rotate_copy_paste}  
\end{figure}

The proposed VRCrop strategy is depicted in Fig.~\ref{fig:rotate_copy_paste}(b). The four steps are described as follows: 

\textbf{Step 1:} All points of the rare instance from a random point cloud are copied. For example, in Fig.~\ref{fig:rotate_copy_paste}(b), the points $\left\{\boldsymbol{p}_{22}, \boldsymbol{p}_{23}, \boldsymbol{p}_{32}, \boldsymbol{p}_{33}\right\}$ in the ``Point Cloud B" is copied. 

\textbf{Step 2:} The copied points are rotated by a rotation matrix $R$, which is expressed by the Eq.~(\ref{eq:rotate_points}),
\begin{equation}\small\label{eq:rotate_points}
	\hat{\boldsymbol{P}} = 
	\left[ \begin{array}{ccc}
		\text{cos}\left(\phi\right)   & -\text{sin}\left(\phi\right) & 0 \\
		\text{sin}\left(\phi\right)   & \text{cos}\left(\phi\right)  & 0 \\
		0                               & 0                              & 1
	\end{array} \right] \times \boldsymbol{P},
\end{equation}
where $\boldsymbol{P} \in \mathbb{R}^{3\times M}$ is the concatenation of the $M$ points; $\hat{\boldsymbol{P}}$ indicates the output points; $\phi$ means the angle of rotation. Note that by Eq.~(\ref{eq:rotate_points}), the horizontal coordinate $u$ of each rotated point is changed, so all $\boldsymbol{u}$ values of the points after rotation should be updated again by the Alg.~\ref{alg:scan_unfolding++}. In the example, for ease of description, we use $\left\{\hat{\boldsymbol{p}}_{22}, \hat{\boldsymbol{p}}_{23}, \hat{\boldsymbol{p}}_{32}, \hat{\boldsymbol{p}}_{33}\right\}$ to indicate the rotated points. Besides, the $M$ in Fig.~\ref{fig:rotate_copy_paste}(b) is set to 4.

\textbf{Step 3:} Conflicted points in the current point cloud are detected and deleted. For example, according to the $\left(\boldsymbol{u}, \boldsymbol{v}\right)$ coordinates of the rotated points $\left\{\hat{\boldsymbol{p}}_{22}, \hat{\boldsymbol{p}}_{23}, \hat{\boldsymbol{p}}_{32}, \hat{\boldsymbol{p}}_{33}\right\}$, VRCrop strategy checks whether there exist points with the same $\left(\boldsymbol{u}, \boldsymbol{v}\right)$ coordinates in ``Point Cloud A". If there are some points with the same $\left(\boldsymbol{u}, \boldsymbol{v}\right)$ coordinates, VRCrop strategy deletes them in ``Point Cloud A". 

\textbf{Step 4:} The rotated points are pasted to the current point cloud. For example, VRCrop strategy pastes the rotated points $\left\{\hat{\boldsymbol{p}}_{22}, \hat{\boldsymbol{p}}_{23}, \hat{\boldsymbol{p}}_{32}, \hat{\boldsymbol{p}}_{33}\right\}$ to ``Point Cloud A". 

In addition, we here provide the conventional copy-rotate-paste or copy-paste operation for comparison. Specifically, existing approaches like PolarMix~\cite{polarmix_2022} directly append the rotated points to the current point cloud (see Fig.~\ref{fig:rotate_copy_paste}(a)). The conventional strategy causes a substantial increase in the number of points in the current point cloud after data augmentation. Moreover, the conventional strategy brings undesirable artifacts to the current point cloud, thereby bringing difficulty in training the range image-point fused models.

\section{Experiments}
In this section, we first describe experimental settings. Then, we discuss the importance of the proposed virtual range image-guided copy-rotate-paste (VRCrop) strategy. Next, we validate the effectiveness of the proposed trainable pointwise decoder module (PDM). Finally, we provide more experimental results.    

\subsection{Experimental Settings}
\textbf{Dataset.} All experiments are conducted on SemanticKITTI \cite{semantickitti_2019_behley}, SemanticPOSS~\cite{semanticposs_2020}, and nuScenes~\cite{nuscenes_panoptic} datasets. SemanticKITTI is a large-scale outdoor point cloud dataset, which provides high-quality per-point labels. In the dataset, sequences $\left\{00 \sim 07, 09 \sim 10\right\}$, $\left\{08\right\}$, and $\left\{11 \sim 21\right\}$ are used as the training, validation, and test sets, respectively. Besides, only 19 classes are considered under the single scan condition. SemanticPOSS includes six sequences $\left\{00\sim05\right\}$ where the sequence $\left\{02\right\}$ serves as the test dataset. Moreover, 14 classes are considered in this dataset. The nuScenes contains 28,130 training, 6,019 validation, and 6,008 test point clouds, respectively. Additionally, 16 annotated classes are considered.

\textbf{Models and Implementation Details.} The range image-based models, namely RangeNet53++~\cite{rangenet++}, FIDNet~\cite{fidnet_2021}, CENet~\cite{cenet_2022}, and Fast FMVNet~\cite{filling_missing2024}, serve as the baselines because they can execute efficiently. Besides, all models are trained for 50, 50, and 80 epochs on the SemanticKITTI, SemanticPOSS, and nuScenes datasets, respectively. The optimizer AdamW is used to train the models with a learning rate of 0.002. During the training and testing phases, all random seeds are fixed to ``123" for reproduction. Moreover, the intersection-over-union (IoU) score over each class and mean IoU (mIoU) score over all classes serve as the evaluation metrics, \textit{i.e.}, ``IoU = $\frac{TP}{TP + FP + FN}$" where TP, FP, and FN means true positive, false positive, and false negative predictions.

\subsection{With or Without VRCrop Strategy?}
In this section, we show why the VRCrop strategy in data augmentation is important for training range image-based models with the proposed trainable pointwise decoder module (PDM).

\begin{figure}[t]
	\centering
	\includegraphics[width=0.98\columnwidth]{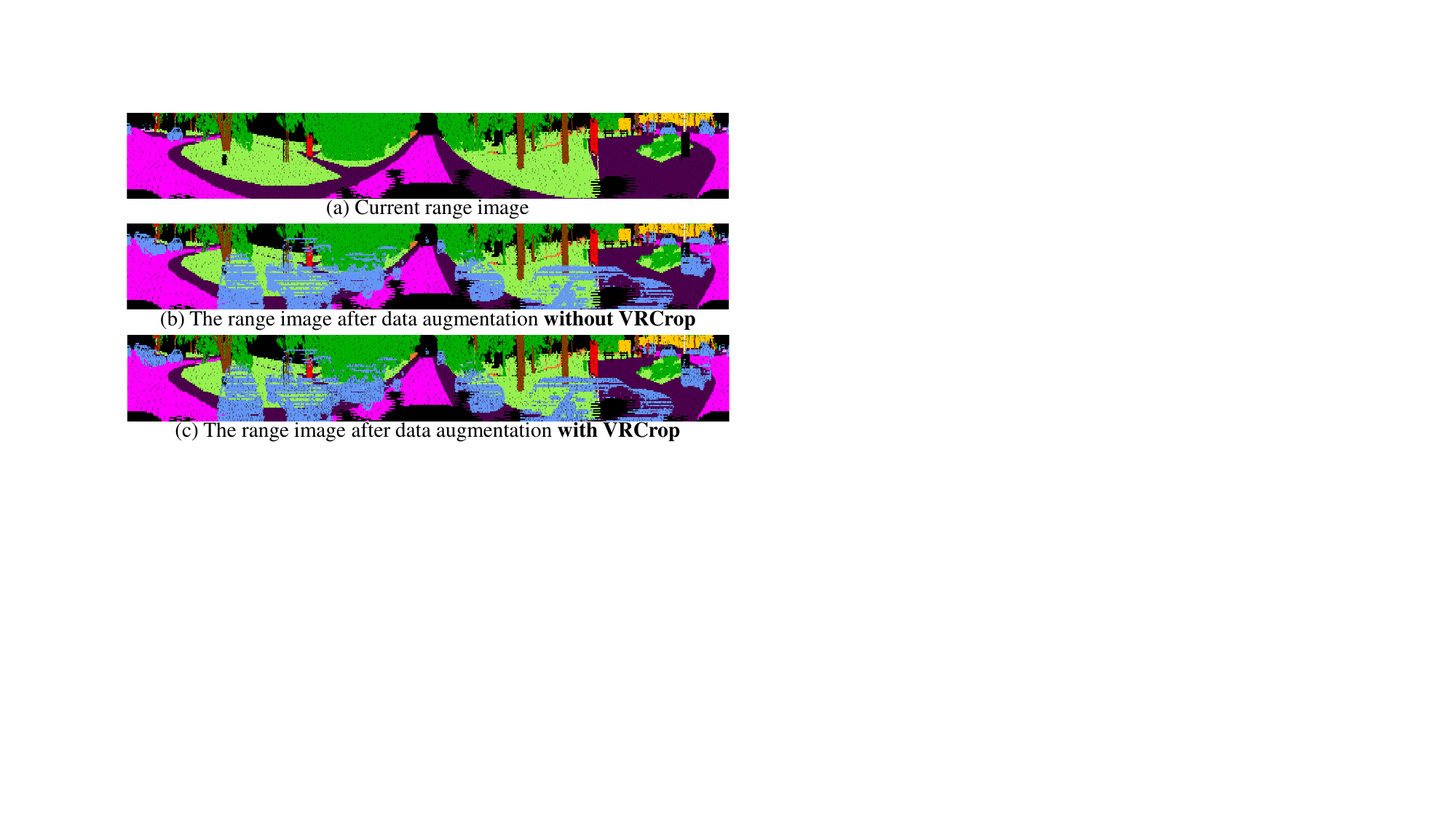}
	\caption{(a) is a range image without applying any copy-rotate-paste operation for comparison. (b) and (c) are the range images augmented without and with the proposed VRCrop strategy, respectively. Note that for ease of description, only \textit{cars} serve as the rare instances in (b) and (c).}
	\label{fig:vrcrop_range_img}  
\end{figure}

\begin{table}[t]
	\caption{Comparison results among range image-based models trained with or without the proposed VRCrop strategy in terms of mIoU scores (\%). ``\ding{52}": with the proposed VRCrop strategy.}
	\label{tab:vrcrop_range_img}
	\centering
	\scalebox{1.0}{
		\begin{tabular}{l|c|c}
			\hline
			Modules                     & VRCrop    & mIoU  \\ \hline \hline 
			\multirow{2}*{RangeNet53++} &           & 64.41 \\ \cline{2-3}
			& \ding{52} & \textcolor{blue}{64.41} \\ \hline \hline
			
			\multirow{2}*{FIDNet}       &           & 65.09 \\ \cline{2-3}
			& \ding{52} & \textcolor{blue}{65.97} \\ \hline \hline
			
			\multirow{2}*{CENet}        &           & 66.05 \\ \cline{2-3}
			& \ding{52} & \textcolor{blue}{66.32} \\ \hline \hline
			
			\multirow{2}*{Fast FMVNet}  &           & 67.34 \\ \cline{2-3}
			& \ding{52} & \textcolor{blue}{67.45} \\ \hline 
	\end{tabular}}
\end{table}

\subsubsection{Impacts on Range Image-based Models} 
We train all range image-based models with and without the proposed VRCrop to provide fair comparison results for the following experiments. The models are trained on the SemanticKITTI training dataset, and the results are reported on the validation dataset (Table~\ref{tab:vrcrop_range_img}).  

In Table~\ref{tab:vrcrop_range_img}, we see that RangeNet53++ trained with VRCrop achieves the same performance as the counterpart trained without VRCrop. Other models trained with VRCrop obtain slightly better performance than their counterparts. Therefore, for training the range image-based models, the proposed VRCrop strategy should be considered, although the range images prepared with and without VRCrop look very similar (see (b) and (c) in Fig.\ref{fig:vrcrop_range_img}). Note that for ease of description, only \textit{cars} serve as rare instances in Fig.\ref{fig:vrcrop_range_img}.

\subsubsection{Impacts on the Models with PDM}
We first provide qualitative comparisons on the point clouds after augmentation with and without the proposed VRCrop strategy. Then, we show how many points need to be processed in the point clouds after augmentation with and without the proposed VRCrop. Finally, we demonstrate the positive impacts of VRCrop on the range image-point fused models, \textit{i.e.}, RangeNet53+PDM, FIDNet+PDM, CENet+PDM, and Fast FMVNet+PDM. All models are trained on the SemanticKITTI training dataset, and the results are reported on the validation dataset.   

\begin{figure}[t]
	\centering
	\includegraphics[width=1.0\columnwidth]{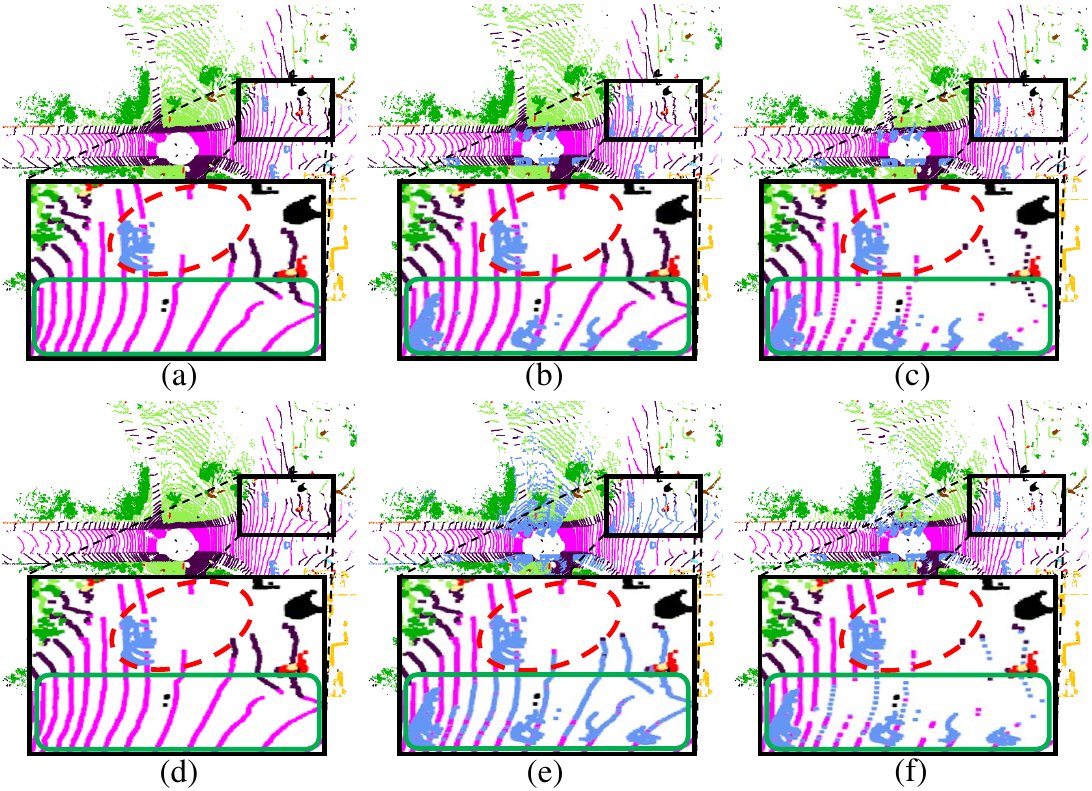}
	\caption{(a) is the ground truth point cloud without any data augmentation for comparison. (b) and (c) are the ground truth point clouds after data augmentation without and with the proposed VRCrop strategy, respectively. (d), (e), and (f) are the point clouds produced by the following two steps: projecting (a), (b), and (c) point clouds onto the range image and then re-projecting the points from the range image back onto the point clouds. The dashed red ellipses highlight that there are no points in the area behind the ``original" \textit{car} (\textit{i.e.}, all cars are indicated by \textcolor[RGB]{96,152,249}{blue color points}). Some areas after applying data augmentation in (b), (c), (e), and (f) are emphasized by the green rectangles. \textbf{Best viewed in color and zoom in}.}
	\label{fig:vrcrop_artifacts}  
\end{figure}

\textbf{Undesirable Artifacts.} As shown in Fig.~\ref{fig:vrcrop_artifacts}, the ``original" car blocks the LiDAR beams, and hence the region behind the ``original" car has no points (\textit{i.e.}, see the regions in Fig.~\ref{fig:vrcrop_artifacts} highlighted by the dashed red ellipses). Fig.~\ref{fig:vrcrop_artifacts}(b) shows the ground truth point cloud after data augmentation without VRCrop. However, we see that the region behind the newly ``pasted" cars is filled with many points, as emphasized by the green rectangle in Fig.~\ref{fig:vrcrop_artifacts}(b). This situation violates the principle of the LiDAR sensor and introduces undesirable artifacts behind these ``pasted" cars. By contrast, Fig.~\ref{fig:vrcrop_artifacts}(c) shows only few points behind the ``pasted" cars in the same region. Almost no artifacts are brought. 

More importantly, the data augmentation without VRCrop has an adverse impact on ``range image-to-points projection" outputs. During the training phase, the neighbor points and corresponding feature vectors are re-projected from the range image back onto the point cloud for the operations in PDM (see Sec.~\ref{sec:trainable_pdm}). Here, for ease of description, we first project the ground truth point cloud onto the range image, and then the ground truth points in the range image are projected back onto the point cloud (see (d), (e), and (f) in Fig.~\ref{fig:vrcrop_artifacts}). Fig.~\ref{fig:vrcrop_artifacts}(e) shows that the points behind these ``pasted" cars should be classified as the car class (see the blue points in the green rectangle), but the corresponding ground truth points in Fig.~\ref{fig:vrcrop_artifacts}(b) belong to the road class (see the pink points in the green rectangle). The mismatched information between Fig.~\ref{fig:vrcrop_artifacts}(b) and Fig.~\ref{fig:vrcrop_artifacts}(e) is caused by the unwanted artifacts behind the ``pasted" cars. The mismatched semantics bring difficulty in training the models with the proposed PDM, and then decrease the segmentation performance (see Table~\ref{tab:vrcrop_pdm_model}). By contrast, with the proposed VRCrop in data augmentation, the ground truth point cloud in Fig.~\ref{fig:vrcrop_artifacts}(c) and the re-projected points in Fig.~\ref{fig:vrcrop_artifacts}(f) are almost identical and without mismatched semantics. Therefore, the qualitative comparison results validate the importance of the proposed VRCrop. 

\begin{table}[t]
	\caption{Comparison results between the data augmentation without and with the proposed VRCrop strategy. ``Max": the maximal number of increased points on the point cloud after data augmentation. ``Align": the maximal number of points set for alignment during training. ``-": no values.}
	\label{tab:vrcrop_increased_points}
	\centering
	\scalebox{0.935}{
		\begin{tabular}{l|c|c|c|c|c}
			\hline
			\multirow{2}{*}{Datasets} & \multirow{2}{*}{Virtual Sizes} & \multicolumn{2}{c|}{Without VRCrop} & \multicolumn{2}{c}{With VRCrop} \\ \cline{3-6}
			&                & Max    & Align   & Max  & Align    \\ \hline \hline
			\multirow{2}{*}{SemanticKITTI}  & \multirow{2}{*}{$64\times2180$} & \multirow{2}{*}{40,907} & \multirow{2}{*}{170,299} &  \multirow{2}{*}{-} & 139,520 \\ 
			&                &        &         &      & (\textcolor{blue}{-30,779}) \\ \hline
			
			\multirow{2}{*}{SemanticPOSS}  & \multirow{2}{*}{$40\times1800$} & \multirow{2}{*}{14,786} & \multirow{2}{*}{85,672}  &  \multirow{2}{*}{-}  & 72,000 \\ 
			&                &        &         &      & (\textcolor{blue}{-13,672}) \\ \hline
			
			\multirow{2}{*}{nuScenes} & \multirow{2}{*}{$32\times1090$} & \multirow{2}{*}{19,901} & \multirow{2}{*}{54,781} & \multirow{2}{*}{-} & 34,880   \\ 
			&     &        &         &      & (\textcolor{blue}{-19,901}) \\ \hline
	\end{tabular}}
\end{table}

\textbf{The Maximal Number of Points in Alignment.} During the training phase, the maximal number of points should be aligned for parallel training. Also, the total number of points in the point cloud is related to the training time and max allocated GPU memory. Comparison results about the total numbers of points in the point clouds after augmentation without and with VRCrop are provided in Table~\ref{tab:vrcrop_increased_points}. 

Note that for ease of description, the rare instances here in the SemanticKITTI~\cite{semantickitti_2019_behley}, SemanticPOSS~\cite{semanticposs_2020}, and nuScenes~\cite{nuscenes_panoptic} datasets are only pasted to the current point cloud once. Besides, in this paper, the classes of the rare instances in the three datasets are listed as follows: 
\begin{itemize}
	\item SemanticKITTI: \textit{Bicycle}, \textit{Motorcycle}, \textit{Truck}, \textit{Other-vehicle}, \textit{Person}, \textit{Bicyclist}, \textit{Motorcyclist}, \textit{Other-ground}, \textit{Trunk}, \textit{Pole}, and \textit{Traffic-sign}. 
	
	\item SemanticPOSS: \textit{Rider}, \textit{Trunk}, \textit{Traffic-sign}, \textit{Pole}, \textit{Trashcan}, \textit{Cone/stone}, \textit{Fence}, and \textit{Bike}.
	
	\item nuScenes: \textit{Barrier}, \textit{Bicycle}, \textit{Bus}, \textit{Car}, \textit{Construction Vehicle}, \textit{Motorcycle}, \textit{Pedestrian}, \textit{traffic Cone}, \textit{Trailer}, and \textit{Truck}.
\end{itemize}
In addition, for the data augmentation without VRCrop, the maximal number of points in alignment is the summation of ``the maximal number of points of all rare instances in a point cloud" and ``the maximal number of points in a point cloud before data augmentation". For example, in SemanticKITTI, the maximal number of points of all rare instances in a point cloud is 40,907. The maximal number of points in a point cloud before data augmentation is 129,392. Hence, the maximal number of points in the alignment for parallel training is set to 170,299 ($40,907 + 129,392 = 170,299$). By contrast, for the data augmentation with the proposed VRCrop, the maximal number of points in alignment is the size of the virtual range image, \textit{e.g.}, 139,520 ($64\times2180$) for SemanticKITTI. The rest of the experimental results on three datasets are provided in Table~\ref{tab:vrcrop_increased_points}. 

We see that without the proposed VRCrop in data augmentation, the numbers of points in alignment are increased dramatically, \textit{i.e.}, 30,779 ($170,299 - 139,520$) for SemanticKITTI, 13,672 ($85,672 - 72,000$) for SemanticPOSS, and 19,901 ($54,781 - 34,880$) for nuScenes. More importantly, the increased points lead to increased training time, as depicted in the ``Time" column in Table~\ref{tab:vrcrop_pdm_model}. Also, the increased points result in increased max allocated GPU memory during training (see the ``Mem." column in Table~\ref{tab:vrcrop_pdm_model}).

\begin{table}[t]
	\caption{Comparison results among models trained with or without the proposed VRCrop strategy in terms of max allocated GPU memory (Mem.) during training, training time (unit: hour), and mIoU scores (\%). Here, the window size and Top-$K$ for PDM are set to $5\times5$ and 7, respectively. ``\ding{52}": with the proposed VRCrop strategy. }
	\label{tab:vrcrop_pdm_model}
	\centering
	\scalebox{1.0}{
		\begin{tabular}{l|c|c|c|c}
			\hline
			Modules                  & VRCrop    & Mem.    & Time   & mIoU  \\ \hline \hline 
			\multirow{2}*{RangeNet53+PDM} &  & 18.07GB & 23.75h & 62.91 \\ \cline{2-5}
			& \ding{52} & 13.12GB & 19.63h & \textcolor{blue}{64.84} \\ \hline \hline
			
			\multirow{2}*{FIDNet+PDM}&           & 21.33GB & 26.10h & 65.06 \\ \cline{2-5}
			& \ding{52} & 16.38GB & 20.98h & \textcolor{blue}{66.23} \\ \hline \hline
			
			\multirow{2}*{CENet+PDM} &           & 18.21GB & 24.92h & 64.64 \\ \cline{2-5}
			& \ding{52} & 13.27GB & 20.38h & \textcolor{blue}{66.33} \\ \hline \hline
			
			\multirow{2}*{Fast FMVNet+PDM} & & 20.71GB & 26.17h & 66.03 \\ \cline{2-5}
			& \ding{52} & 16.15GB & 22.32h & \textcolor{blue}{68.55} \\ \hline 
	\end{tabular}}
\end{table}

\textbf{Decreased Model Performance.} The description above show that during training, the undesirable artifacts decrease the segmentation performance, and the increased number of points leads to the raised training time and max allocated GPU memory. The quantitative comparisons among the range image-point fused models trained with and without VRCrop are provided in Table~\ref{tab:vrcrop_pdm_model}. Note that during training, the rare instances here are rotated and pasted to the current point cloud three times.  

In Table~\ref{tab:vrcrop_pdm_model}, we see that RangeNet53+PDM trained without and with VRCrop obtains 62.91\% and 64.84\% (+1.93) mIoU scores, respectively. Similarly, FIDNet+PDM achieves 65.06\% and 66.23\% (+1.17) mIoU scores, respectively. CENet+PDM gets 64.64\% and 66.33\% (+1.69) mIoU scores, respectively. Fast FMVNet+PDM obtains 66.03\% and 68.55\% (+2.52) mIoU scores, respectively. As explained before, the undesirable artifacts introduced by the data augmentation without the proposed VRCrop bring difficulty in training the models, thereby leading to decreased segmentation performance. In addition, we see that without VRCrop, the corresponding training time and max allocated GPU memory are significantly increased due to the dramatic increase in the total number of points after augmentation. Therefore, we validate the importance of the proposed VRCrop strategy in training range image-based models with the trainable pointwise decoder module.

\subsection{Effectiveness of Pointwise Decoder Module}\label{sec:pdm_effectiveness}
In this subsection, we first validate the effectiveness of the proposed local feature extraction module by comparing it with alternatives. Then, we provide parameter analysis results about the window size and top-$K$ in the proposed range image-guided $K$NN search. Note that we here only use Fast FMVNet+PDM to conduct experiments due to the best performance among the four models. 

\subsubsection{Local Feature Extraction Module}
We here provide various options to the designed local feature extraction component in the proposed trainable pointwise decoder module (PDM) for comparison. They are described as follows:

\textbf{Voting-A.} Inspired by the voting strategy, we directly compute the average (AVG) of all features of the neighbor points as the output of the query point $\boldsymbol{p}_i$. The formula is expressed by $\boldsymbol{o}_i = \text{AVG}\left\{\boldsymbol{f}_j | j \in \mathcal{N}(i)\right\}$ where the $\mathcal{N}(i)$ means a local region formed by the point $\boldsymbol{p}_i$. Also, a classifier is appended to make a prediction. For ease of description, we name this option ``Voting-A".

\textbf{Voting-B.} Similarly, Voting-B concatenates all features of neighbor points as the output. The formula is expressed by $\boldsymbol{o}_i = \text{Concate}\left\{\boldsymbol{f}_j | j \in \mathcal{N}(i)\right\}$. Besides, in the following classifier, the kernel size of a convolution is set to $K \times 1$. Voting-B learns a linear combination of the features of the neighbor points.

\textbf{Pointwise-MLP-A.} It is the same as the design in PointNet++~\cite{pointnet++_2017}. In a local region $\mathcal{N}(i)$ formed by the point $\boldsymbol{p}_i$, each neighbor point feature vector $\boldsymbol{f}_j$ is first concatenated with the relative position $\triangle\boldsymbol{p}_{ij}$. Then, the output goes through an MLP layer. Subsequently, a max pooling operation (MAX) is applied to all neighbors to produce the feature vector $\boldsymbol{o}_i$. This is expressed by the following formula $\boldsymbol{o}_i = \text{MAX}\left\{\text{MLP}\left(\text{Concate}\left\{\triangle\boldsymbol{p}_{ij},\boldsymbol{f}_j\right\}
\right) | j \in \mathcal{N}(i)\right\}$. Finally, a classifier is used to make a prediction. For ease of description, we name this option as ``Pointwise-MLP-A".

\textbf{Pointwise-MLP-B.} Same as the EdgeConv in DGCNN \cite{dgcnn2019}, Pointwise-MLP-B uses the features $\left\{\triangle\boldsymbol{f}_{ij}, \boldsymbol{f}_i\right\}$ instead of $\left\{\triangle\boldsymbol{p}_{ij},\boldsymbol{f}_j\right\}$ in Pointwise-MLP-A to build the relationships between the query point $\boldsymbol{p}_i$ and the neighbor points. The formula is described by the expression $\boldsymbol{o}_i = \text{MAX}\left\{\text{MLP}\left(\text{Concate}\left\{\triangle\boldsymbol{f}_{ij}, \boldsymbol{f}_i\right\}
\right) | j \in \mathcal{N}(i)\right\}$. Also, a classifier is adopted to make a prediction.

\textbf{Attentive-Weight-A.} Similar to the design in PCNN \cite{pcnn2018}, Attentive-Weight-A first computes the weights on relative positions $\triangle\boldsymbol{p}_{ij}$. Then, the weights are multiplied by the neighbor point features $\boldsymbol{f}_j$. Finally, a summation operation (SUM) is adopted to aggregate features of all neighbor points. The expression is provided by the formula $\boldsymbol{o}_i = \text{SUM}\left\{\text{Softmax}\left(\text{MLP}
\left(\triangle\boldsymbol{p}_{ij}\right)\right) \odot \boldsymbol{f}_j | j \in \mathcal{N}(i)\right\}$. Finally, a classifier is appended to make a prediction. For ease of description, we name this as ``Attentive-Weight-A".

\textbf{Attentive-Weight-B.} Instead of the summation operation in Attentive-Weight-A, the concatenation of the weighted features is adopted here. This is expressed by $\boldsymbol{o}_i = \text{Concate}\left\{\text{Softmax}\left(\text{MLP}
\left(\triangle\boldsymbol{p}_{ij}\right)\right) \odot \boldsymbol{f}_j | j \in \mathcal{N}(i)\right\}$. Moreover, same as the setting in Voting-B, the kernel size of convolution in the following classifier is set to $K\times1$. This helps to learn a linear combination of the features of the neighbor points.

\textbf{Attentive-Weight-C.} Here, the ``Scaled Dot-Product Attention"~\cite{attention2017} is utilized to aggregate local features from all neighbor points. Specifically, the query point feature vector $\boldsymbol{f}_i$ is multiplied by each neighbor point feature vector $\boldsymbol{f}_j$. Then, the output is divided by $\sqrt{d}$, and a softmax function is used to get weights. Finally, the weighted features are calculated as the output. The formula is described by the expression $\boldsymbol{o}_i = \text{Softmax}\left(\frac{\boldsymbol{f}_i^{T} \boldsymbol{F}}{\sqrt{d}}\right) \boldsymbol{F}^T$ where $\boldsymbol{F} = \left[\boldsymbol{f}_1, \dots, \boldsymbol{f}_K\right] \in \mathbb{R}^{d \times K}$ is the concatenation of the features of the $K$ neighbor points. Finally, a classifier is used to make a prediction. In this option, the attentive weights are derived from the relationships between the features of the query point and the features of the neighbor points.

\textbf{Attentive-Weight-D.} Different from the Attentive-Weight-C above, Attentive-Weight-D builds the relationships directly between the query point and its neighbor points. The calculated weights are then applied to the features of the $K$ neighbor points. The expression is presented by the formula $\boldsymbol{o}_i = 
\text{Softmax}\left(\frac{\text{MLP}(\boldsymbol{p}_i)^{T}\text{MLP}(\boldsymbol{P})}{\sqrt{h}}\right)\boldsymbol{F}^T$  
where $\boldsymbol{P} = [\boldsymbol{p}_1, \dots, \boldsymbol{p}_K] \in \mathbb{R}^{3 \times K}$ is a collection of all neighbor points; And $\boldsymbol{F} = \left[\boldsymbol{f}_1, \dots, \boldsymbol{f}_K\right] \in \mathbb{R}^{d \times K}$ is a set of features of the neighbor points; And $h$ means the dimension of the transformed features $\text{MLP}\left(\boldsymbol{p}_i\right) \in \mathbb{R}^{h}$.

\begin{table}[t]
	\caption{Comparison results among various local feature extraction modules in terms of parameters (Param.), FLOPs, frames per second (FPS), and mIoU scores (\%). Here, the window size and Top-$K$ are set to $5\times5$ and 7, respectively.}
	\label{tab:local_feature_extraction}
	\centering
	\scalebox{1.0}{
		\begin{tabular}{l|c|c|c|c}
			\hline
			Modules         & Param. & FLOPs   & FPS   & mIoU  \\ \hline \hline 
			Voting-A        & 4.35M  & 195.53G & 32.61 & 66.84 \\ \hline
			Voting-B        & 4.36M  & 197.05G & 32.58 & 67.55 \\ \hline \hline
			Pointwise-MLP-A & 4.35M  & 199.35G & 28.91 & 67.57 \\ \hline
			Pointwise-MLP-B & 4.35M  & 200.95G & 29.34 & 67.84 \\ \hline \hline
			Attentive-Weight-A&4.35M & 197.58G & 29.17 & 67.16 \\ \hline
			Attentive-Weight-B&4.36M & 199.09G & 28.99 & 66.79 \\ \hline
			Attentive-Weight-C&4.35M & 195.59G & 31.07 & 67.28 \\ \hline
			Attentive-Weight-D&4.35M & 196.24G & 28.05 & 67.63 \\ \hline \hline
			Ours              &4.35M & 201.23G & 26.53 & \textbf{68.55} \\ \hline
	\end{tabular}}
\end{table}

\textbf{Analysis of Local Feature Extraction Modules.} The experimental results are provided in Table~\ref{tab:local_feature_extraction}. Note that during the training phase, all models have the same experimental settings, such as the $5\times5$ window size and Top-$K$ of 7. The models are trained on the SemanticKITTI~\cite{semantickitti_2019_behley} training dataset, and the results are reported on the validation dataset. 

In Table~\ref{tab:local_feature_extraction}, we see that the model with our module achieves the best performance (68.55\% mIoU score) among all models with the local feature extraction modules and still keeps a high execution speed (26.53 FPS). This is because using both edge features and relative positions to calculate the weights is more effective compared with others. Besides, the combination of relative position information and the features of the neighbor points is beneficial for segmentation performance. 

By contrast, Voting-A and Voting-B only adopt the features of the neighbor points to make a prediction. Pointwise-MLP-A and Pointwise-MLP-B merely consider how to build distinguishing local features but ignore the calculation of weights to select these features effectively. For Attentive-Weight-A/B/C/D, only utilizing the relative positions, the features of the points, and the points to compute weights is inferior to ours. Also, only adopting the features of the neighbor points to calculate the output is sub-optimal. The impressive segmentation performance in Table~\ref{tab:comparison_existing_ppm} validate the effectiveness of the proposed local feature extraction module in PDM.

\begin{table}[t]
	\caption{Comparison results among the models with different window sizes in terms of parameters (Param.), FLOPs, frames per second (FPS), and mIoU scores (\%). Here, Top-$K$ is set to 7.}
	\label{tab:window_sizes}
	\centering
	\scalebox{1.0}{
		\begin{tabular}{c|c|c|c|c}
			\hline
			Sizes      & Param. & FLOPs   & FPS   & mIoU  \\ \hline \hline 
			$3\times3$ & 4.35M  & 201.23G & 28.87 & 68.04 \\ \hline
			$5\times5$ & 4.35M  & 201.23G & 26.53 & \textbf{68.55} \\ \hline
			$7\times7$ & 4.35M  & 201.23G & 23.27 & 68.47 \\ \hline
			$9\times9$ & 4.35M  & 201.23G & 20.08 & 67.96 \\ \hline
	\end{tabular}}
\end{table}

\begin{table}[t]
	\caption{Comparison results among the models with various top-$K$ values in terms of parameters (Param.), FLOPs, frames per second (FPS), and mIoU scores (\%). The window size is set to $5\times5$.}
	\label{tab:topk_7}
	\centering
	\scalebox{1.0}{
		\begin{tabular}{c|c|c|c|c}
			\hline
			Top-$K$   & Param. & FLOPs   & FPS   & mIoU  \\ \hline \hline 
			3         & 4.35M  & 197.97G & 29.57 & 67.21 \\ \hline
			5         & 4.35M  & 199.60G & 27.93 & 67.91 \\ \hline
			7         & 4.35M  & 201.23G & 26.53 & \textbf{68.55} \\ \hline
			9         & 4.35M  & 202.85G & 25.08 & 68.13 \\ \hline
	\end{tabular}}
\end{table}

\subsubsection{Parameter Analysis of the Window Size}
The window size decides how many neighbor points are initially considered in the range image-guided $K$NN search in the proposed trainable pointwise decoder module (see Sec.~\ref{sec:knn_search}). The parameter analysis results on the SemanticKITTI validation dataset are provided in Table~\ref{tab:window_sizes}. Note that the default Top-$K$ is set to 7.

In Table~\ref{tab:window_sizes}, we see that the model with the window size of $5\times5$ obtains the best performance (68.55\% mIoU score). Besides, a large window size such as $9\times9$ results in inferior performance because the large size introduces much semantically unrelated information from neighbor points. Moreover, the large window size decreases the model speed significantly. In this paper, we choose the window size of $5\times5$ considering the balance between the model performance and execution speed.

\subsubsection{Parameter Analysis of the Top-$K$}
The top-$K$ decides how many neighbor points are finally selected for the downstream local feature extraction module (see Sec.~\ref{sec:knn_search}). Experimental results on the SemanticKITTI validation dataset are reported in Table~\ref{tab:topk_7}. Note that the window size is set to $5\times5$.

In Table~\ref{tab:topk_7}, we see that by selecting the top 7 nearest neighbor points in the proposed trainable pointwise decoder module, the model obtains the best performance among all options. By comparison, choosing fewer neighbor points (\textit{e.g.}, top-$K$ of 3) decreases the model performance significantly, from 68.55\% to 67.21\%. Here, we choose the top-$K$ of 7 due to the consideration of the speed-accuracy trade-off.

\begin{table*}[t]
	\caption{Quantitative comparisons on the SemanticKITTI test set in terms of IoU and mIoU scores (\%). ``$\dagger$" indicates that \textbf{TTA} is applied to the results. Note that \textbf{NO TTA} is applied to our results. The best result is emphasized by the \textbf{bold} font. The second best result is highlighted by the \underline{underline}.}
	\label{tab:64x2048_kitti_test_results}
	\centering
	\scalebox{0.78}{
		\begin{tabular}{l|c|l|c|c|c|c|c|c|c|c|c|c|c|c|c|c|c|c|c|c|c}
			\hline
			Models	& Years & mIoU &\rotatebox{90}{Car} &\rotatebox{90}{Bicycle} &\rotatebox{90}{Motorcycle} &\rotatebox{90}{Truck} &\rotatebox{90}{Other-vehicle} &\rotatebox{90}{Person} &\rotatebox{90}{Bicyclist} &\rotatebox{90}{Motorcyclist} &\rotatebox{90}{Road} &\rotatebox{90}{Parking} &\rotatebox{90}{Sidewalk} &\rotatebox{90}{Other-ground} &\rotatebox{90}{Building} &\rotatebox{90}{Fence} &\rotatebox{90}{Vegetation} &\rotatebox{90}{Trunk} &\rotatebox{90}{Terrain} &\rotatebox{90}{Pole} &\rotatebox{90}{Traffic-sign}  \\ \hline \hline
			
			SqueezeSeg~\cite{squeezeseg} & 2018 &30.8 &68.3 &18.1 &5.1 &4.1 &4.8 &16.5 &17.3 &1.2 &84.9 &28.4 &54.7 &4.6 &61.5 &29.2 &59.6 &25.5 &54.7 &11.2 &36.3   \\ \hline
			SqueezeSegV2~\cite{squeezesegv2} & 2019 &39.7 &81.8 &18.5 &17.9 &13.4 &14.0 &20.1 &25.1 &3.9 &88.6 &45.8 &67.6 &17.7 &73.7 &41.1 &71.8 &35.8 &60.2 &20.2 &36.3  \\ \hline
			
			RangeNet21~\cite{rangenet++}& 2019 &47.4 &85.4 &26.2 &26.5 &18.6 &15.6 &31.8 &33.6 &4.0 &91.4 &57.0 &74.0 &26.4 &81.9 &52.3 &77.6 &48.4 &63.6 &36.0 &50.0 \\ \hline 
			
			RangeNet53++~\cite{rangenet++}& 2019 &52.2 &91.4 &25.7 &34.4 &25.7 &23.0 &38.3 &38.8 &4.8 &91.8 &65.0 &75.2 &27.8 &87.4 &58.6 &80.5 &55.1 &64.6 &47.9 &55.9 \\ \hline
			
			SqSegV3-21~\cite{squeezesegv3_2020}& 2020 &51.6 &89.4 &33.7 &34.9 &11.3 &21.5 &42.6 &44.9 &21.2 &90.8 &54.1 &73.3 &23.2 &84.8 &53.6 &80.2 &53.3 &64.5 &46.4 &57.6 \\ \hline
			SqSegV3-53~\cite{squeezesegv3_2020}& 2020 &55.9 &92.5 &38.7 &36.5 &29.6 &33.0 &45.6 &46.2 &20.1 &91.7 &63.4 &74.8 &26.4 &89.0 &59.4 &82.0 &58.7 &65.4 &49.6 &58.9 \\ \hline
			
			FIDNet~\cite{fidnet_2021}& 2021 &59.5 &93.9 &54.7 &48.9 &27.6 &23.9 &62.3 &59.8 &23.7 &90.6 &59.1 &75.8 &26.7 &88.9 &60.5 &84.5 &64.4 &69.0 &53.3 &62.8  \\ \hline
			
			CENet$\dagger$~\cite{cenet_2022}& 2022 &64.7 &91.9 &58.6 &50.3 &40.6 &42.3 &68.9 &65.9 &43.5 &90.3 &60.9 &75.1 &31.5 &91.0 &66.2 &84.5 &69.7 &70.0 &61.5 &\underline{67.6} \\ \hline 
			
			RangeViT~\cite{rangevit_2023} & 2023 & 64.0 & 95.4 &55.8 &43.5 &29.8 &42.1 &63.9 &58.2 &38.1 &\textbf{93.1} &70.2 &\textbf{80.0} &32.5 &92.0 &69.0 &\underline{85.3} &70.6 &\underline{71.2} &60.8 &64.7 \\ \hline
			
			RangeFormer~\cite{rangeformer_2023}& 2023 &\underline{\textcolor{blue}{69.5}} &\textcolor{blue}{94.7} &\textcolor{blue}{60.0} &\underline{\textcolor{blue}{69.7}} &\underline{\textcolor{blue}{57.9}} &\underline{\textcolor{blue}{64.1}} &\textcolor{blue}{72.3} &\underline{\textcolor{blue}{72.5}} &\underline{\textcolor{blue}{54.9}} &\textcolor{blue}{90.3} &\textcolor{blue}{69.9} &\textcolor{blue}{74.9} &\underline{\textcolor{blue}{38.9}} &\textcolor{blue}{90.2} &\textcolor{blue}{66.1} &\textcolor{blue}{84.1} &\textcolor{blue}{68.1} &\textcolor{blue}{70.0} &\textcolor{blue}{58.9} &\textcolor{blue}{63.1} \\ \hline
			RangeFormer$\dagger$~\cite{rangeformer_2023}& 2023 &\textbf{73.3} &\textbf{96.7} &\textbf{69.4} &\textbf{73.7} &\textbf{59.9} &\textbf{66.2} &\textbf{78.1} &\textbf{75.9} &\textbf{58.1} &92.4 &\underline{73.0} &78.8 &\textbf{42.4} &92.3 &\textbf{70.1} &\textbf{86.6} &\textbf{73.3} &\textbf{72.8} &\textbf{66.4} &66.6 \\ \hline

			FMVNet~\cite{filling_missing2024} &2024 &68.0 &\underline{96.6} & 63.4 & 60.9 & 42.1 & 55.5 &\underline{75.6} & 70.7 & 26.1 & 92.5 &\textbf{73.8} &\underline{79.3} & 37.7 &\underline{92.3} &\underline{69.3} & 85.2 &\underline{71.4} & 69.7 &\underline{63.0} & 66.8 \\ \hline
			
			Fast FMVNetv2 (Ours) &2024 &\textcolor{blue}{68.1} &\textcolor{blue}{96.1} &\underline{\textcolor{blue}{64.2}} &\textcolor{blue}{66.4} &\textcolor{blue}{42.3} & \textcolor{blue}{49.1} & \textcolor{blue}{74.9} & \textcolor{blue}{67.9} & \textcolor{blue}{43.5} &\underline{\textcolor{blue}{92.6}} & \textcolor{blue}{72.0} & \textcolor{blue}{79.1} & \textcolor{blue}{34.0} &\textbf{\textcolor{blue}{92.5}} & \textcolor{blue}{68.5} & \textcolor{blue}{84.5} & \textcolor{blue}{68.6} & \textcolor{blue}{69.5} & \textcolor{blue}{59.1} &\textbf{\textcolor{blue}{68.5}} \\ \hline	
	\end{tabular}}
\end{table*}

\begin{table*}[t]
	\caption{Quantitative comparison results on the SemanticPOSS test dataset (sequence $\left\{02\right\}$) in terms of IoU and mIoU scores (\%). Note that \textbf{NO TTA} is employed in our results. ``$\ast$": the model pre-trained on the Cityscapes~\cite{cityscapes16} dataset. The best result is emphasized by the \textbf{bold} font. The second best result is highlighted by the \underline{underline}.}
	\label{tab:64x2048_poss_test_results}
	\centering
	\scalebox{0.86}{
		\begin{tabular}{l|l|c|c|c|c|c|c|c|c|c|c|c|c|c}
			\hline
			Models &mIoU &People &Rider &Car &Trunk &Plants &Traffic Sign &Pole &Trashcan &Building &Cone/Stone &Fence &Bike &Ground \\ \hline \hline
			SqueezeSeg~\cite{squeezeseg}     &18.9 &14.2 &1.0  &13.2 &10.4 &28.0 &5.1  &5.7  &2.3  &43.6 &0.2  &15.6 &31.0 &75.0 \\ \hline 
			SqueezeSegV2~\cite{squeezesegv2} &30.0 &48.0 &9.4  &48.5 &11.3 &50.1 &6.7  &6.2  &14.8 &60.4 &5.2  &22.1 &36.1 &71.3 \\ \hline
			MINet~\cite{minet_2021}          &43.2 &62.4 &12.1 &63.8 &22.3 &68.6 &16.7 &30.1 &28.9 &75.1 &28.6 &32.2 &44.9 &76.3 \\ \hline 
			
			RangeNet53++~\cite{rangenet++,filling_missing2024} &51.4 &74.6 &22.6 &79.8 &\textbf{26.9} &71.3 &21.3 &28.2 &31.6 &77.5 &49.3 &51.7 &54.9 &77.9 \\ \hline 
			FIDNet~\cite{fidnet_2021,filling_missing2024}      &53.5 &78.5 &29.6 &79.0 &25.8 &71.4 &23.3 &32.8 &38.4 &79.2 &49.4 &54.4 &55.9 &78.2 \\ \hline
			CENet~\cite{cenet_2022,filling_missing2024}        &54.3 &78.1 &29.0 &83.0 &26.4 &70.5 &22.9 &\underline{33.6} &36.6 &79.2 &\textbf{58.1} &53.1 &56.2 &79.6 \\ \hline 
			
			FMVNet~\cite{filling_missing2024} &54.4 &78.7 &\underline{30.2} &80.7 &24.5 &73.2 &\textbf{26.0} &\textbf{35.0} &35.6 &82.8 &53.5 &51.5 &56.6 &79.5 \\ \hline
			FMVNet$^\ast$~\cite{filling_missing2024} &\underline{55.1} &80.0 &29.9 &\underline{84.2} &26.2 &\underline{73.4} &\underline{25.5} &31.4 &34.9 &82.5 &\underline{55.0} &\textbf{55.9} &56.4 &\underline{80.9} \\ \hline 
			
			Fast FMVNet~\cite{filling_missing2024} &\textcolor{blue}{54.3} &78.7 &27.3 &82.6 &26.6 &73.1 &25.4 &32.4 &\underline{39.0} &81.7 &45.8 &54.9 &\underline{57.6} &80.3 \\ \hline
			Fast FMVNet$\ast$~\cite{filling_missing2024} &\textcolor{blue}{54.7} &80.1 &29.2 &83.9 &\underline{26.7} &73.1 &24.8 &32.7 &\textbf{40.8} &81.4 &48.8 &54.8 &56.3 &78.4 \\ \hline \hline
			
			Fast FMVNetv2 (Ours) &\textcolor{blue}{54.4} &\underline{80.1} &\textbf{30.8} &83.5 &24.9 &72.5 &24.1 &31.9 &33.2 &\textbf{82.9} &50.7 &55.3 &\textbf{57.8} &79.7 \\ \hline
			Fast FMVNetv2$^\ast$(Ours)&\textcolor{blue}{\textbf{55.3}} &\textbf{80.2} &30.0 &\textbf{85.0} &25.8 &\textbf{73.6} &23.7 &31.6 &37.9 &\underline{82.8} &54.9 &\underline{55.7} &57.1 &\textbf{81.1} \\ \hline
	\end{tabular}}
\end{table*}

\begin{table}[htp]
	\caption{Comparison results among the models with different post-processing (Post-p.) approaches in terms of parameters (Param.), FLOPs, frames per second (FPS), and mIoU scores (\%).}
	\label{tab:comparison_existing_ppm}
	\centering
	\scalebox{1.0}{
		\begin{tabular}{l|c|c|c|c|c}
			\hline
			Models                      & Post-p.  & Param. & FLOPs   & FPS   & mIoU  \\ \hline \hline 
			\multirow{3}*{RangeNet53}   & $K$NN    & 50.38M & 360.22G & 63.83 & 64.41 \\ \cline{2-6}
			& KPConv   & 50.46M & 370.49G & 36.50 & 64.54 \\ \cline{2-6}
			& PDM      & 50.40M & 366.82G & 32.78 & \textcolor{blue}{64.84} \\ \hline \hline
			
			\multirow{3}*{FIDNet}       & NLA      & 6.05M  & 339.39G & 60.18 & 65.97 \\ \cline{2-6}
			& KPConv   & 6.19M  & 357.31G & 34.18 & 65.81 \\ \cline{2-6}
			& PDM      & 6.10M  & 350.02G & 30.99 & \textcolor{blue}{66.23} \\ \hline \hline
			
			\multirow{3}*{CENet}        & NLA      & 6.77M  & 434.02G & 62.71 & 66.32 \\ \cline{2-6}
			& KPConv   & 6.91M  & 451.94G & 34.73 & 65.87 \\ \cline{2-6}
			& PDM      & 6.82M  & 444.64G & 31.45 & \textcolor{blue}{66.33} \\ \hline \hline
			
			\multirow{3}*{Fast FMVNet}  & NLA      & 4.31M  & 190.60G & 47.10 & 67.45 \\ \cline{2-6}
			& KPConv   & 4.44M  & 208.52G & 28.75 & 67.72 \\ \cline{2-6}
			& PDM      & 4.35M  & 201.23G & 26.53 & \textcolor{blue}{68.55} \\ \hline               
	\end{tabular}}
\end{table}

\subsection{Comparison with Post-processing Approaches}
In this section, we compare the proposed trainable pointwise decoder module (PDM) with existing post-processing approaches, namely $K$NN~\cite{rangenet++}, NLA~\cite{fidnet_2021}, and KPConv~\cite{kpconv_2019,rangevit_2023}. In experiments, the window sizes for $K$NN~\cite{rangenet++} and NLA~\cite{fidnet_2021} are set to $7\times7$, and the corresponding top-$K$ values are set to 7. For PDM, the window size and top~$K$ are set to $5\times5$ and 7, respectively. For KPConv, the kernel size and the number of neighbors are set to 15 and 7, respectively. The experimental results are reported in Table~\ref{tab:comparison_existing_ppm}.

We see that by employing the proposed PDM, all models achieve the best performance, \textit{i.e.}, 64.84\% for RangeNet53+PDM, 66.23\% for FIDNet+PDM, 66.33\% for CENet+PDM, and 68.55\% for Fast FMVNet+PDM. Besides, the models with PDM still maintain high speeds, \textit{i.e.}, 32.78 FPS for RangeNet53+PDM, 30.99 FPS for FIDNet+PDM, 31.45 FPS for CENet+PDM, and 26.53 FPS for Fast FMVNet+PDM. By comparison, the models with KPConv maintain slightly higher speeds but achieve inferior segmentation performance. Similarly, the four models with $K$NN or NLA obtain low mIoU scores, although they achieve higher speeds. Note that $K$NN and NLA have used the proposed range image-guided $K$NN search, so they can run very fast. The experimental results validate the effectiveness of the proposed PDM.

\subsection{More Performance Comparison}
In this subsection, we first provide more comparison results on the SemanticKITTI, SemanticPOSS, and nuScenes datasets. Then, we show time comparison results among various models. Note that for ease of description, we name Fast FMVNet+PDM as Fast FMVNetv2. Besides, we only adopt Fast FMVNetv2 in the following experiments due to its impressive segmentation performance. 

\begin{table*}[h]
	\caption{Quantitative comparison results on the nuScenes validation dataset in terms of IoU and mIoU scores (\%). Note that \textbf{NO TTA} is applied to our results. ``Constr. Veh.": ``Construction Vehicle"; ``Drive. Sur.": ``Driveable Surface"; STR: scalable training from range view strategy~\cite{rangeformer_2023}; $\ddagger$: the model pre-trained on ImageNet~\cite{imagenet}; $\ast$: the model pre-trained on Cityscapes~\cite{cityscapes16}. The best result is emphasized by the \textbf{bold} font. The second best result is highlighted by the \underline{underline}.}
	\label{tab:32x1088_nuscenes_val_results}
	\centering
	\scalebox{0.93}{
		\begin{tabular}{l|l|c|c|c|c|c|c|c|c|c|c|c|c|c|c|c|c}
			\hline
			Models	 &mIoU &\rotatebox{90}{Barrier} &\rotatebox{90}{Bicycle} &\rotatebox{90}{Bus} &\rotatebox{90}{Car} &\rotatebox{90}{Constr. Veh.} &\rotatebox{90}{Motorcycle} &\rotatebox{90}{Pedestrian} &\rotatebox{90}{Traffic Cone} &\rotatebox{90}{Trailer} &\rotatebox{90}{Truck} &\rotatebox{90}{Drive. Sur.} &\rotatebox{90}{Other Flat} &\rotatebox{90}{Sidewalk} &\rotatebox{90}{Terrain} &\rotatebox{90}{Manmade} &\rotatebox{90}{Vegetation}  \\ \hline \hline
			RangeNet53++~\cite{rangenet++,nuscenes_panoptic} &65.5 &66.0 &21.3 &77.2  &80.9 &30.2 &66.8 &69.6 &52.1 &54.2 &72.3 &94.1 &66.6 &63.5 &70.1 &83.1 &79.8 \\ \hline
			RangeNet53++~\cite{rangenet++,filling_missing2024} &71.1 &58.5 &38.1 &90.0  &84.0 &46.1 &80.1 &62.3 &42.3 &62.4 &80.9 &96.5 &73.7 &75.1 &74.2 &87.6 &86.0 \\ \hline
			FIDNet~\cite{fidnet_2021,filling_missing2024} &73.5 &59.5 &44.2 &88.4 &84.6 &48.1 &84.0 &70.4 &59.9 &65.7 &78.0 &96.5 &71.6 &74.7 &75.1 &88.7 &87.3 \\ \hline 
			CENet~\cite{cenet_2022,filling_missing2024}  &73.4 &60.2 &43.0 &88.0 &85.0 &53.6 &70.4 &71.0 &62.5 &65.6 &80.1 &96.6 &72.3 &74.9 &75.1 &89.1 &87.7 \\ \hline
			
			RangeViT$\ddagger$~\cite{rangevit_2023} &74.8 &75.1 &39.0 &90.2 &88.4 &48.0 &79.2 &77.2 &66.4 &65.1 &76.7 &96.3 &71.1 &73.7 &73.9 &88.9 &87.1 \\ \hline 
			RangeViT$^\ast$~\cite{rangevit_2023} &75.2 &75.5 &40.7 &88.3 &90.1 &49.3 &79.3 &77.2 &66.3 &65.2 &80.0 &96.4 &71.4 &73.8 &73.8 &89.9 &87.2 \\ \hline 
			
			RangeFormer+STR$^\ast$~\cite{rangeformer_2023}&\underline{77.1} &\underline{76.0} &44.7 &94.2 &\underline{92.2} &54.2 &82.1 &76.7 &\textbf{69.3} &61.8 &83.4 &96.7 &\textbf{75.7} &75.2 &75.4 &88.8 &87.3 \\ \hline
			RangeFormer$^\ast$~\cite{rangeformer_2023}&\textbf{78.1} &\textbf{78.0} &45.2 &94.0 &\textbf{92.9} &58.7 &83.9 &\underline{77.9} &69.1 &63.7 &\underline{85.6} &96.7 &74.5 &75.1 &75.3 &89.1 &87.5 \\ \hline 
			
			FMVNet~\cite{filling_missing2024} &76.7 &61.5 &\textbf{50.0} &94.7 &86.9 &\underline{59.0} &\textbf{87.3} &\textbf{78.0} &54.4 &\underline{69.1} &85.1 &\textbf{97.0} &74.1 &\underline{76.3} &\underline{75.7} &\underline{90.2} &\underline{88.7} \\ \hline 
			FMVNet$^\ast$~\cite{filling_missing2024}&76.8 &61.1 &\underline{49.5} &94.7 &86.8 &\textbf{59.6} &71.1 &77.2 &\underline{69.1} &\textbf{70.9} &\textbf{85.6} &96.9 &\underline{75.0} &\textbf{76.5} &\textbf{75.8} &90.1 &88.4 \\ \hline 
			
			Fast FMVNet~\cite{filling_missing2024} &\textcolor{blue}{75.6} &60.3 &45.4 &89.9 &86.6 &55.1 &85.3 &75.1 &64.2 &67.0 &84.6 &96.7 &72.0 &75.0 &74.5 &89.5 &87.9 \\ \hline 
			Fast FMVNet$^\ast$~\cite{filling_missing2024}&\textcolor{blue}{76.0} &60.3 &45.8 &95.1 &86.7 &54.7 &85.7 &74.0 &66.2 &67.1 &83.5 &96.7 &72.7 &75.1 &74.8 &89.8 &88.3 \\ \hline \hline
			
			Fast FMVNetv2 (Ours) &\textcolor{blue}{75.8} &64.4 &42.9 &\underline{95.2} &87.2 &56.2 &\underline{86.9} &70.8 &62.2 &66.6 &83.8 &96.7 &72.3 &75.1 &74.2 &90.1 &88.4 \\ \hline 
			Fast FMVNetv2$^\ast$(Ours)&\textcolor{blue}{76.1} &64.8 &41.0 &\textbf{95.3} &88.6 &54.5 &84.9 &72.5 &62.2 &67.1 &85.5 &\underline{96.9} &74.8 &75.5 &75.1 &\textbf{90.4} &\textbf{88.8} \\ \hline  
	\end{tabular}}
\end{table*}

\subsubsection{Comparison Results on SemanticKITTI}
The comparison results on the SemanticKITTI test dataset are reported in Table~\ref{tab:64x2048_kitti_test_results}. In experiments, the pre-trained Fast FMVNet~\cite{filling_missing2024} on the Cityscapes~\cite{cityscapes16} dataset is adopted to initialize our Fast FMVNetv2. Then, we fine-tune Fast FMVNetv2 on the training and validation datasets for 50 epochs and get the results from the benchmark. Note that here we do not adopt any test-time augmentation (TTA) techniques further to improve the model performance for a fair comparison. 

In Table~\ref{tab:64x2048_kitti_test_results}, we see that without TTA, our Fast FMVNetv2 achieves a higher mIoU score than the recent works RangeViT~\cite{rangevit_2023} and FMVNet~\cite{filling_missing2024}. Moreover, the overall performance of our Fast FMVNetv2 is inferior to that of RangeFormer~\cite{rangeformer_2023}, but Fast FMVNetv2 surpasses RangeFormer over 12 classes (19 classes in total), namely \textit{Car}, \textit{Bicycle}, \textit{Person}, \textit{Road}, \textit{Parking}, \textit{Sidewalk}, \textit{Building}, \textit{Fence}, \textit{Vegetation}, \textit{Trunk}, \textit{Pole}, and \textit{Traffic sign}. Besides, our Fast FMVNetv2 can run at a higher speed than RangeFormer (see Table~\ref{tab:time_models} and Fig.~\ref{fig:speed_miou}). The experimental results show that our Fast FMVNetv2 obtains a better speed-accuracy trade-off compared with other models. Also, the results validate the effectiveness of the proposed approaches.

\subsubsection{Comparison Results on SemanticPOSS} 
The comparison results on the SemanticPOSS test dataset (sequence $\left\{02\right\}$) are reported in Table~\ref{tab:64x2048_poss_test_results}. In experiments, Fast FMVNetv2 is trained from scratch. Fast FMVNetv2$^\ast$ is first pre-trained on the Cityscapes dataset~\cite{cityscapes16} and then fine-tuned on the SemanticPOSS training dataset. Besides, for a fair comparison, we do not employ any TTA techniques to improve the segmentation performance further. 

In Table~\ref{tab:64x2048_poss_test_results}, we see that Fast FMVNetv2 and Fast FMVNetv2$^\ast$ consistently achieve better performance than their counterparts. Specifically, Fast FMVNetv2 gets the 54.4\% mIoU score while Fast FMVNet obtains the 54.3\% mIoU score. By adopting pre-trained weights on the Cityscapes dataset during training, Fast FMVNetv2$^\ast$ achieves the 55.3\% mIoU score which is higher than the 54.7\% mIoU score of Fast FMVNet$^\ast$. The experimental results validate the effectiveness of the proposed trainable pointwise decoder module.

\subsubsection{Comparison Results on nuScenes}
The comparison results on the nuScenes validation dataset are provided in Table~\ref{tab:32x1088_nuscenes_val_results}. Similar to the experiments above, Fast FMVNetv2 is trained on the nuScenes training dataset from scratch, and Fast FMVNetv2$^\ast$ is trained with the pre-trained model on the Cityscapes dataset. Besides, no TTA is applied to improve our results further. In Table~\ref{tab:32x1088_nuscenes_val_results}, we see that our Fast FMVNetv2 and Fast FMVNetv2$^\ast$ consistently obtain better performance than the counterparts, namely Fast FMVNet and Fast FMVNet$^\ast$ (\textit{i.e.}, 75.8\% vs. 75.6\% mIoU scores, and 76.1\% vs. 76.0\% mIoU scores). The improved performance validates the effectiveness of the proposed trainable pointwise decoder module. 

Besides, our Fast FMVNetv2$^\ast$ is inferior to RangeFormer$^\ast$. This might be caused by the erroneously labelled points in the nuScenes dataset~\cite{lasermix_2023,filling_missing2024}. During training, same as the work~\cite{filling_missing2024}, we remove erroneously annotated points and corresponding labels in training our models on the nuScenes training dataset. However, we consider the erroneously annotated points on the nuScenes validation dataset during the testing phase for fair comparisons. Due to at least 5.7\% erroneously labelled points in the validation dataset, our Fast FMVNetv2$^\ast$ achieves sub-optimal performance.  

\begin{table}[t]
	\caption{Time comparison results among the models in terms of the number of model parameters (Params.), latency, frames per second (FPS), and mIoU scores (\%) on the SemanticKITTI~\cite{semantickitti_2019_behley} validation dataset. ``$\ast$": models pre-trained on the Cityscapes dataset.}
	\label{tab:time_models}
	\centering
	\scalebox{0.93}{
		\begin{tabular}{l|c|c|c|c|c}
			\hline
			Methods                                & Years &Params. & Latency &FPS   & mIoU  \\ \hline  \hline
			MinkowskiNet~\cite{minkowski2019}      & 2019  &21.7M   & 48.4ms  &20.7  & 61.1  \\ \hline  
			Cylinder3D~\cite{cylindrical3d2021}    & 2021  &56.3M   & 71.5ms  &13.3  & 65.9  \\ \hline  
			UniSeg 0.2$\times$~\cite{openpcseg2023}& 2023  &28.8M   & 84.6ms  &11.8  & 67.0  \\ \hline 
			UniSeg 1.0$\times$~\cite{openpcseg2023}& 2023  &147.6M  & 145.0ms &6.9   & 71.3  \\ \hline \hline
			
			RangeNet53+PDM~\cite{rangenet++}       & 2019  &50.4M   & 30.5ms  &32.8  & 64.8  \\ \hline
			FIDNet+PDM~\cite{fidnet_2021}          & 2021  &6.1M    & 32.3ms  &31.0  & 66.2  \\ \hline
			CENet+PDM~\cite{cenet_2022}            & 2022  &6.8M    & 31.8ms  &31.5  & 66.3  \\ \hline
			RangeFormer$^\ast$~\cite{rangeformer_2023,filling_missing2024} & 2023  &24.3M   & 90.3ms  &11.1  & 67.6  \\ \hline \hline
			
			Fast FMVNet$^\ast$~\cite{filling_missing2024} & 2024  &4.3M    & 20.8ms  &48.1  & 67.9  \\ \hline 
			Fast FMVNetv2$^\ast$ (Ours)                   & 2024  &4.4M    & 37.7ms  &26.5  & 68.6 \\ \hline
	\end{tabular}}
\end{table}

\subsubsection{Time Comparison}
\begin{figure}[t]
	\centering
	\includegraphics[width=0.75\columnwidth]{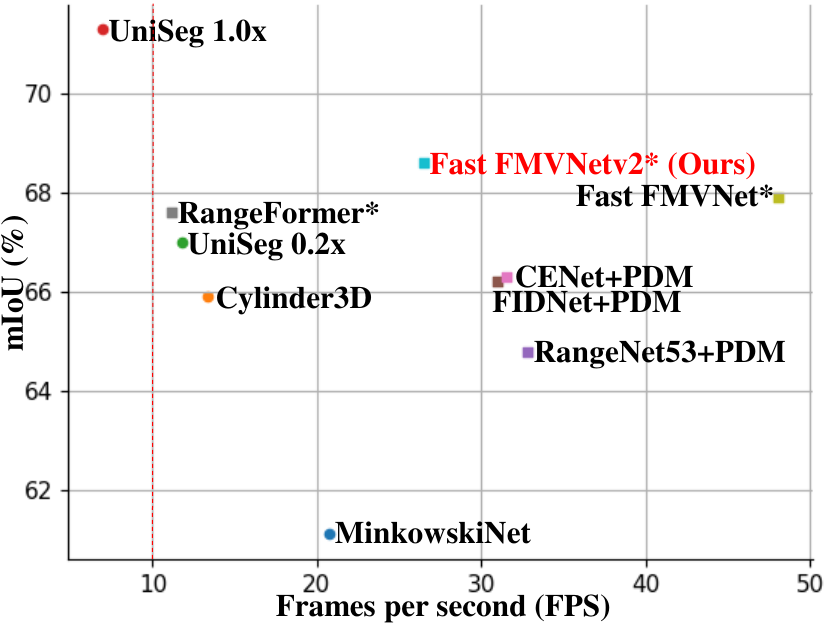}
	\caption{Comparison results among different models in terms of frames per second (FPS) and mIoU scores (\%).}
	\label{fig:speed_miou}
\end{figure}

To claim the advantage of the proposed approaches, we here provide time comparison results in Table~\ref{tab:time_models} and Fig.~\ref{fig:speed_miou}. For RangeFormer, we utilize the results from the work~\cite{filling_missing2024} for a fair comparison. Also, we adopt RangeNet53+PDM, FIDNet+PDM, and CENet+PDM for comparisons because they achieve better performance than their counterparts in terms of mIoU scores. Besides, we use the results of Fast FMVNetv2$^\ast$ which is pre-trained on the Cityscapes~\cite{cityscapes16} dataset. Moreover, all models are tested on an NVIDIA A100 GPU.

Table~\ref{tab:time_models} shows that Fast FMVNetv2$^\ast$ achieves a higher mIoU score than the counterpart Fast FMVNet$^\ast$. Besides, Fast FMVNetv2$^\ast$ not only gets a higher mIoU score than that of RangeFormer$^\ast$~\cite{rangeformer_2023,filling_missing2024} but also runs at a higher speed, \textit{i.e.}, about two times faster than RangeFormer$^\ast$. Moreover, note that our Fast FMVNetv2$^\ast$ only has 4.4M model parameters.
The comparison results in Table~\ref{tab:time_models} and Fig.~\ref{fig:speed_miou} show that Fast FMVNetv2$^\ast$ is lightweight and achieves a better speed-accuracy trade-off than other models.

\section{Conclusion}
Point cloud segmentation plays a critical role in the robot perception task. In this paper, we introduced a trainable pointwise decoder module (PDM) as a post-processing method. PDM can be trained with the range image-based segmentation models in an end-to-end manner and process varying-density outdoor point clouds well, thereby enabling the models to achieve better performance than their counterparts. Also, the models with PDM are still efficient. In addition, we proposed a virtual range image-guided copy-rotate-paste (VRCrop) strategy in the point cloud data augmentation for training range image-point fused models. During training, VRCrop can restrict the total number of points in the point cloud after data augmentation so as to avoid a high computational cost and long training time. More importantly, VRCrop alleviates the problem of occurring undesirable artifacts in the augmented point cloud. The experimental results on the SemanticKITTI, SemanticPOSS, and nuScenes datasets have validated the effectiveness of the proposed PDM and VRCrop. However, VRCrop cannot completely remove the points behind the newly ``pasted" instances. A clustering approach can be integrated into VRCrop to overcome this problem.

%\section*{Acknowledgments}
%This should be a simple paragraph before the References to thank those individuals and institutions who have supported your work on this article.

\bibliographystyle{IEEEtran}
\bibliography{main}

\clearpage

{\appendix
	The content in the supplementary material is summarized as follows:
	
	\begin{itemize}
		\item More implementation details are provided.
		
		\item More comparison results on the SemanticKITTI validation dataset are provided. 
	\end{itemize}

	\section*{Implementation Details}\label{sec:more_imp_details}
	All models are trained for 50, 50, and 80 epochs on the SemanticKITTI, SemanticPOSS, and nuScenes datasets, respectively. Besides, the optimizer AdamW is used to train all models with a learning rate of 0.002. The weight decay is set to 0.0001. A warm-up learning rate scheduler is adopted where a cosine function serves as the warm-up strategy, and an exponential decay is utilized. For training models on the SemanticKITTI and SemanticPOSS datasets, the numbers of warm-up epochs are set to 10. For nuScenes, the number is set to 20. 
	
	The data augmentation techniques are listed as follows: (1) ``random scaling" with a probability of 0.5 and the scaling factor within $\left[0.95, 1.05\right]$; (2) ``random flipping" with a probability of 0.5; (3) ``random rotation" with a probability of 0.9; (4) ``RangeMix"~\cite{rangeformer_2023} with a probability of 0.9; (5) ``copy-rotate-paste"~\cite{polarmix_2022} with a probability of 0.9. (6) ``$K$NNI"~\cite{filling_missing2024}. 
	
	In addition, for the experimental results in Table~\ref{tab:vrcrop_range_img}, the batch size for RangeNet53++, FIDNet, and CENet is set to 16. The batch size for Fast FMVNet is set to 8. The batch size settings are consistent with that in the work~\cite{filling_missing2024}. Therefore, the comparisons in Table~\ref{tab:vrcrop_range_img} are fair. For the results in Table~\ref{tab:64x2048_poss_test_results}, the batch size for training Fast FMVNetv2 and Fast FMVNetv2$^\ast$ on the SemanticPOSS dataset is set to 2 due to the limited number of point clouds. For the rest results, we train our models with a batch size of 8 to avoid the out-of-memory issue. 
	
	Besides, note that during the training and testing phases, all random seeds are fixed to ``123" for reproduction. Moreover, no test-time augmentation techniques and the ensemble are applied to our results for fair comparisons.
	
	\section*{Comparison Results on SemanticKITTI}\label{sec:more_results_on_kitti_val}
	As supplements to the experimental results in Table~\ref{tab:comparison_existing_ppm}, the IoU scores over each class on the SemanticKITTI dataset are reported in Table~\ref{tab:64x2048_kitti_val_results}. In addition, the results of Fast FMVNet+PDM$^\ast$ are provided where the model first pre-trained on the Cityscapes dataset and then fine-tuned on the SemanticKITTI training dataset. 
	
	In Table~\ref{tab:64x2048_kitti_val_results}, we see that RangeNet53+PDM, FIDNet+PDM, and Fast FMVNet+PDM consistently achieve better performance than their counterparts. Besides, CENet+PDM obtains similar results to CENet~\cite{filling_missing2024}. The training strategy might cause this. In the paper, we train and test all our models with the fixed random seed ``123" for reproduction. The fixed random seed ``123" may not be the optimal choice for training CENet+PDM. Additionally, with pre-trained weights on the Cityscapes dataset, Fast FMVNet+PDM$^\ast$ surpasses its counterpart Fast FMVNet$^\ast$ in terms of the mIoU score. The experimental results in Table~\ref{tab:64x2048_kitti_val_results} validate the effectiveness of the proposed PDM.

	\begin{table*}[t]
		\caption{Quantitative comparison results on the SemanticKITTI val set in terms of IoU and mIoU scores (\%). Note that \textbf{NO TTA} is applied to our results. ``$\ast$": the model pre-trained on the Cityscapes dataset.}
		\label{tab:64x2048_kitti_val_results}
		\centering
		\scalebox{0.82}{
			\begin{tabular}{l|c|c|c|c|c|c|c|c|c|c|c|c|c|c|c|c|c|c|c|c}
				\hline
				Models	 & mIoU &\rotatebox{90}{Car} &\rotatebox{90}{Bicycle} &\rotatebox{90}{Motorcycle} &\rotatebox{90}{Truck} &\rotatebox{90}{Other-vehicle} &\rotatebox{90}{Person} &\rotatebox{90}{Bicyclist} &\rotatebox{90}{Motorcyclist} &\rotatebox{90}{Road} &\rotatebox{90}{Parking} &\rotatebox{90}{Sidewalk} &\rotatebox{90}{Other-ground} &\rotatebox{90}{Building} &\rotatebox{90}{Fence} &\rotatebox{90}{Vegetation} &\rotatebox{90}{Trunk} &\rotatebox{90}{Terrain} &\rotatebox{90}{Pole} &\rotatebox{90}{Traffic-sign}  \\ \hline \hline
				
				RangeNet53++~\cite{rangenet++,filling_missing2024}&64.4&95.1 &51.6 &72.7 &70.7 &50.2 &75.3 &87.3 &0.0 &95.6 &47.2 &83.0 &14.9 &89.5 &63.3 & 85.8 &64.2 &71.3 &56.9 &48.9 \\ \hline 
				
				RangeNet53+KPConv &64.5&95.9 &49.0 &68.8 &68.5 &54.0 &76.9 &86.5  &0.0 &95.9 &50.1 &83.3 &15.2 &89.4 &62.2 &86.6 &64.3 &72.7 &57.5 &49.7 \\ \hline
				RangeNet53+PDM    &\textcolor{blue}{64.8}&95.7 &50.4 &73.4 &75.3 &50.4 &77.5 &87.7  &0.0 &95.9 &51.1 &83.4 &6.9 &90.2 &63.4 &86.4 &64.5 &72.5 &56.7 &50.7 \\ \hline \hline

				FIDNet~\cite{fidnet_2021,filling_missing2024}&66.0 &94.0 &52.4 &70.0 &76.9 &57.6 &79.3 &85.3&0.0 &95.0 &45.5 &82.5 &20.5 &90.1 &61.1 &87.1 &68.4 &73.3 &63.2 &51.4  \\ \hline
				FIDNet+KPConv &65.8&94.1 &48.3 &68.5 &87.6 &61.5 &80.9 &88.4  &0.6 &94.9 &47.5 &82.6 &9.7 &89.4 &54.5 &87.1 &67.3 &73.1 &62.5 &52.1 \\ \hline
				FIDNet+PDM  &\textcolor{blue}{66.2}&95.0 &53.7 &72.9 &88.1 &59.0 &80.0 &90.3 &0.0  &94.9 &42.1 &82.6 &10.0 &90.0
				&59.3 &86.9 &67.3 &72.7 &61.9 &51.6 \\ \hline \hline

				CENet~\cite{cenet_2022,filling_missing2024}&66.3 &94.2 &49.4 &73.1 &87.6 &59.3 &80.4 &90.0 &0.0 &95.1 &38.8 &81.8 &16.8 &89.4 &58.1 &88.3 &68.0 &75.8 &63.1 &51.0 \\ \hline 
				CENet+KPConv &65.9&94.0 &49.5 &74.0 &81.8 &53.3 &82.3 &92.3 &0.0 &95.2 &42.5 &82.4 &14.7 &90.2 &58.5 &86.9 &68.7 &71.9 &61.2 &52.0 \\ \hline
				CENet+PDM   &\textcolor{blue}{66.3} &94.1 &50.5 &72.0 &83.2 &56.8 &81.8 &91.5 &0.0  &94.9 &41.8 &81.8 &13.5 &90.7 &59.7 &87.7 &70.4 &75.4 &60.4 &54.0 \\ \hline \hline
				
				Fast FMVNet~\cite{filling_missing2024}&67.4&96.1 &50.3 &74.0 &88.6 &67.4 &82.2 &91.1 &0.0 &95.5 &49.2 &83.8 &9.1 &90.6 &63.0 &86.1 &70.4 &70.1 &63.8 &50.2 \\ \hline
				Fast FMVNet+KPConv &67.7&96.4 &48.5 &76.4 &90.0 &65.7 &81.4 &92.1  &0.0 &95.8 &50.3 &83.8 &14.0 &89.9 &60.4 &87.0 &69.0 &72.4 &62.5 &51.1 \\ \hline
				Fast FMVNet+PDM &\textcolor{blue}{68.5} &96.6 &46.5 &76.0 &87.6 &66.5 &81.4 &91.0 &0.0 &95.8 &57.4 &84.1 &16.6 &90.8 &65.6 &87.9 &69.3 &75.6 &61.5 &52.0 \\ \hline \hline
				
				Fast FMVNet$^\ast$~\cite{filling_missing2024} &67.9 &95.3 &52.2 &76.9 &91.6 &52.0 &80.8 &91.7 &0.1 &95.8 &60.7 &84.5 &13.9 &91.2 &65.1 &86.4 &70.2 &70.9 &61.6 &49.3 \\ \hline 
				Fast FMVNet+PDM$^\ast$ &\textcolor{blue}{68.6} &96.8 &51.0 &80.9 &85.8 &68.5 &82.4 &90.8 &0.0 &96.2 &48.1 &84.7 &9.5 &92.2 &68.8 &87.4 &70.8 &72.6 &64.8 &52.2 \\ \hline
		\end{tabular}}
	\end{table*}

	\section*{Qualitative Results on SemanticKITTI}
	We here provide qualitative comparison results among the models trained with and without the proposed VRCrop, and the comparison results among the models with different post-processing approaches. 
	
	\subsection{Models Trained With and Without VRCrop}
	Corresponding to Table~\ref{tab:vrcrop_pdm_model}, the qualitative comparison results among the models trained with and without the proposed VRCrop are provided in Fig.~\ref{fig:vrcrop_no_vrcrop_pdm_comparisons}. We see that the models trained with VRCrop can more accurately predict points than their counterparts. Some differences are emphasized by blue dashed ellipses and rectangles. The low performance for the models trained without VRCrop is caused by undesirable artifacts during training (see Sec.~\ref{sec:vrcrop_strategy}). The experimental results validate the effectiveness of the proposed VRCrop.
	
	\begin{figure*}[t]
		\centering
		\includegraphics[width=1.45\columnwidth]{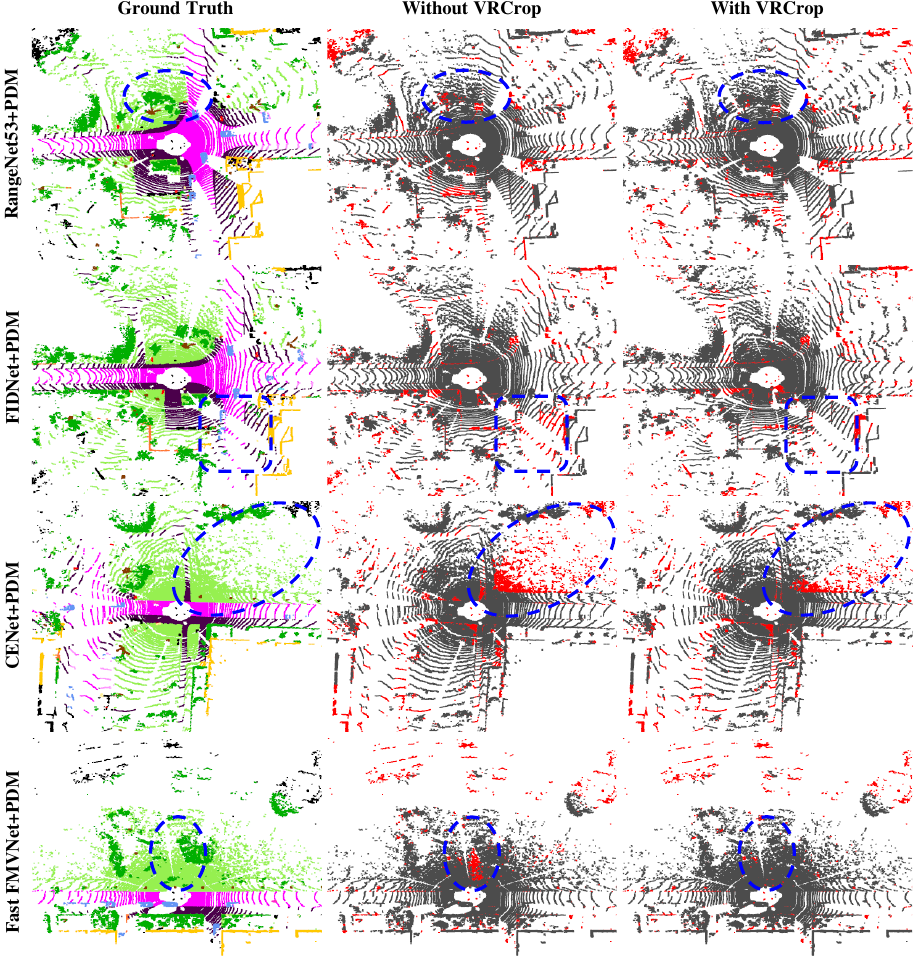}
		\caption{Qualitative comparison results among the models trained with and without the proposed VRCrop. Correct and incorrect predictions are indicated by gray and red colors, respectively. Some differences are highlighted by blue dashed ellipses and rectangles. \textbf{Best viewed in color and zoom-in}.}
		\label{fig:vrcrop_no_vrcrop_pdm_comparisons}
	\end{figure*}
	
	\subsection{Models with Various Post-processing Methods}
	Consistent with the results in Table~\ref{tab:comparison_existing_ppm}, the qualitative comparison results among the models with different post-processing approaches are provided in Fig.~\ref{fig:knnnla_kpconv_pdm_comparisons}. We see that by employing the proposed PDM, the models can accurately predict the points than the counterparts with $K$NN/NLA or KPConv (see the last column in Fig.~\ref{fig:knnnla_kpconv_pdm_comparisons}). The proposed PDM can be trained with the range image-based models in an end-to-end manner. Besides, compared with KPConv, PDM is able to process varying-density outdoor point clouds well. Hence, PDM can improve the performance of the range image-based models. The experiments validate the effectiveness of the proposed PDM.   
	
	\begin{figure*}[t]
		\centering
		\includegraphics[width=1.92\columnwidth]{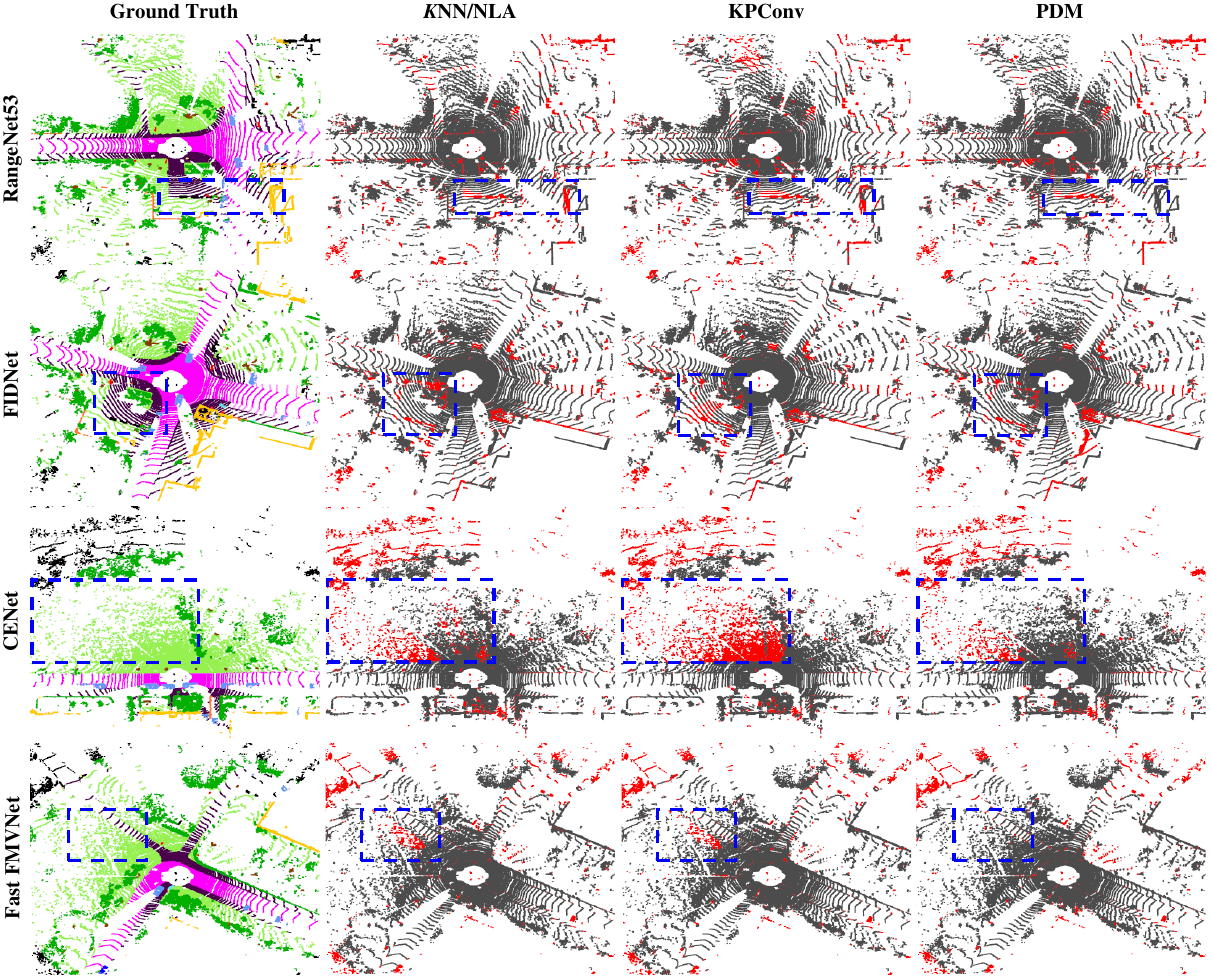}
		\caption{Qualitative comparison results among the models with various post-processing methods. Correct and incorrect predictions are indicated by gray and red colors, respectively. Some differences are highlighted by blue dashed rectangles. \textbf{Best viewed in color and zoom-in}.}
		\label{fig:knnnla_kpconv_pdm_comparisons}
	\end{figure*}
}

\end{document}